\pdfoutput=1

\documentclass[11pt]{article}

\usepackage[preprint]{acl}

\usepackage{times}
\usepackage{latexsym}
\usepackage[T1]{fontenc}
\usepackage[utf8]{inputenc}
\usepackage{microtype}
\usepackage{inconsolata}
\usepackage{graphicx}
\usepackage{svg}
\usepackage{caption}
\usepackage{subcaption}
\usepackage{float}
\usepackage{verbatim}
\usepackage[utf8]{inputenc} % allow utf-8 input
\usepackage[T1]{fontenc}    % use 8-bit T1 fonts
\usepackage{hyperref}       % hyperlinks
\usepackage{url}            % simple URL typesetting
\usepackage{booktabs}       % professional-quality tables
\usepackage{amsfonts}       % blackboard math symbols
\usepackage{nicefrac}       % compact symbols for 1/2, etc.
\usepackage{microtype}      % microtypography
\usepackage[utf8]{inputenc}
\usepackage{color, xcolor, colortbl}
\usepackage{listings}
\usepackage{tabularx}
\usepackage{caption}
\usepackage{geometry}
\definecolor{lightgray}{gray}{0.95}
\definecolor{failred}{RGB}{255, 230, 230}
\usepackage{natbib}
\usepackage{tabularx}
\usepackage{listings}
\usepackage{CJKutf8}
\usepackage{longtable}
\usepackage{color, xcolor, colortbl}
\definecolor{lightgray}{gray}{0.95}
\definecolor{failred}{RGB}{255, 230, 230}
\definecolor{darkgreen}{RGB}{0,100,0} 
\usepackage{amsmath}
\usepackage{multirow} 

\title{Paths Not Taken: Understanding and Mending \\
the Multilingual Factual Recall Pipeline}

\author{
Meng Lu\thanks{Equal contribution.} \\
Brown University \\
\texttt{meng\_lu@brown.edu}
\And
Ruochen Zhang\footnotemark[1] \\
Brown University \\
\texttt{ruochen\_zhang@brown.edu}
\AND
Carsten Eickhoff \\
University of Tübingen \\
\texttt{carsten.eickhoff@uni-tuebingen.de}
\And
Ellie Pavlick \\
Brown University \\
\texttt{ellie\_pavlick@brown.edu}
}

\begin{document}
\begin{CJK*}{UTF8}{gbsn}
\maketitle
\begin{abstract}
Multilingual large language models (LLMs) often exhibit factual inconsistencies across languages, with significantly better performance in factual recall tasks in English than in other languages. The causes of these failures, however, remain poorly understood. Using mechanistic analysis techniques, we uncover the underlying pipeline that LLMs employ, which involves using the English-centric factual recall mechanism to process multilingual queries and then translating English answers back into the target language. We identify two primary sources of error: insufficient engagement of the reliable English-centric mechanism for factual recall, and incorrect translation from English back into the target language for the final answer. To address these vulnerabilities, we introduce two vector interventions, both independent of languages and datasets, to redirect the model toward better internal paths for higher factual consistency. Our interventions combined increase the recall accuracy by over 35 percent for the lowest-performing language. Our findings demonstrate how mechanistic insights can be used to unlock latent multilingual capabilities in LLMs.

\end{abstract}

\section{Introduction}
\label{sec:intro}
Large language models (LLMs) are becoming increasingly multilingual, yet they still demonstrate great language inequalities across various tasks. One issue concerning the reliability of multilingual LLMs is the cross-lingual factual inconsistency ~\citep{Qi_factual_inconsistency_2023}: even though a question like ``\texttt{What is the main religion in Thailand?}'' has only one correct answer, posing it to the same model in different languages can yield conflicting responses. This raises concerns about the reliability of multilingual LLMs given their higher untruthfulness rate when handling non-English inputs \citep{deng2023multilingual,yong2023low,liumultilingualtruthfulness}.

Recent interpretability works suggest that multilingual LLMs ``think'' in their predominant pretraining languages, most often English in the intermediate layers~\citep{wendler2024llamasworkenglishlatent, wu2024semantic}. Follow-up work~\citep{dumas2024llamas, schut2025multilingual} shows that concept and language signals are represented independently. For example, \citet{tang2024language} and~\citet{zhao_handle_multilingualism} observe neurons in the early and late layers in charge of controlling language specificities in the model.

In the context of factual recall, previous studies have mainly focused on English monolingual models~\citep{geva2023dissectingrecallfactualassociations, meng2022locating, chughtai2024summing}, breaking down how facts are stored and retrieved. \citet{constanza_multilingual} and~\citet{wang2025lost} further investigate the process for multilingual models and confirm that intermediate retrieval steps are close to English and answer formation in later layers are language-specific.

These studies together provide converging evidence that across layers, LLMs process inputs in language-specific space, move and solve the task in English-centric concept space, and move back to language-specific output space. However, no existing study has functionally linked these stages into a unified mechanism, nor systematically connected them to specific failure modes underlying cross-lingual factual inconsistencies. To address the gap, we make the following contributions:

\begin{enumerate}
    \item \textbf{Characterizing the multilingual fact recall pipeline}: In Section \ref{sec:pipeline}, we integrate and extend the results from prior work and propose a single hypothesized pipeline that is consistent with model behavior and intervention. Our analysis shows that factual information is first retrieved in English using intermediate English-centric mechanisms, followed by translation into the target language in later model layers.
    
    \item \textbf{Error analysis of multilingual inconsistencies}: By comparing correct and incorrect factual recall instances, we identify two key failure points: (1) the model generates incorrect language-specific answers despite forming correct intermediate English answers, and (2) the model fails to generate correct intermediate English answers in the first place (\S\ref{sec:pipeline}).
    
    \item \textbf{Targeted interventions for error mitigation}: Based on these failure modes, we introduce two language and dataset-independent vector interventions. First, in Section \ref{sec:late-stage}, we leverage the representation difference between recall and translation tasks to promote accurate translation of correct intermediate English answers. For the second, in Section \ref{sec:early-stage}, we derive a general in-context learning signal to enhance the English-centric recall stage.
    
    \item \textbf{Improvement in end-to-end factual recall}: We show that combining both interventions leads to substantial improvements in factual recall—boosting accuracy by up to 37.6 percentage points in the lowest-performing language and achieving an average gain of 19.04 points across all evaluated languages, and outperforming baselines such as explicit translation on held out tasks. 
\end{enumerate}

Together, our work highlights how LLMs can falter when handling information in multilingual contexts. By offering mechanistic insights into the processing pipeline, we identify promising opportunities for targeted interventions that can both uncover latent capabilities and enable more modular control of LLM behaviors.

\section{Multilingual Factual Recall Pipeline}
\label{sec:pipeline}
\begin{figure}[t]
    \centering
\includegraphics[width=\linewidth]{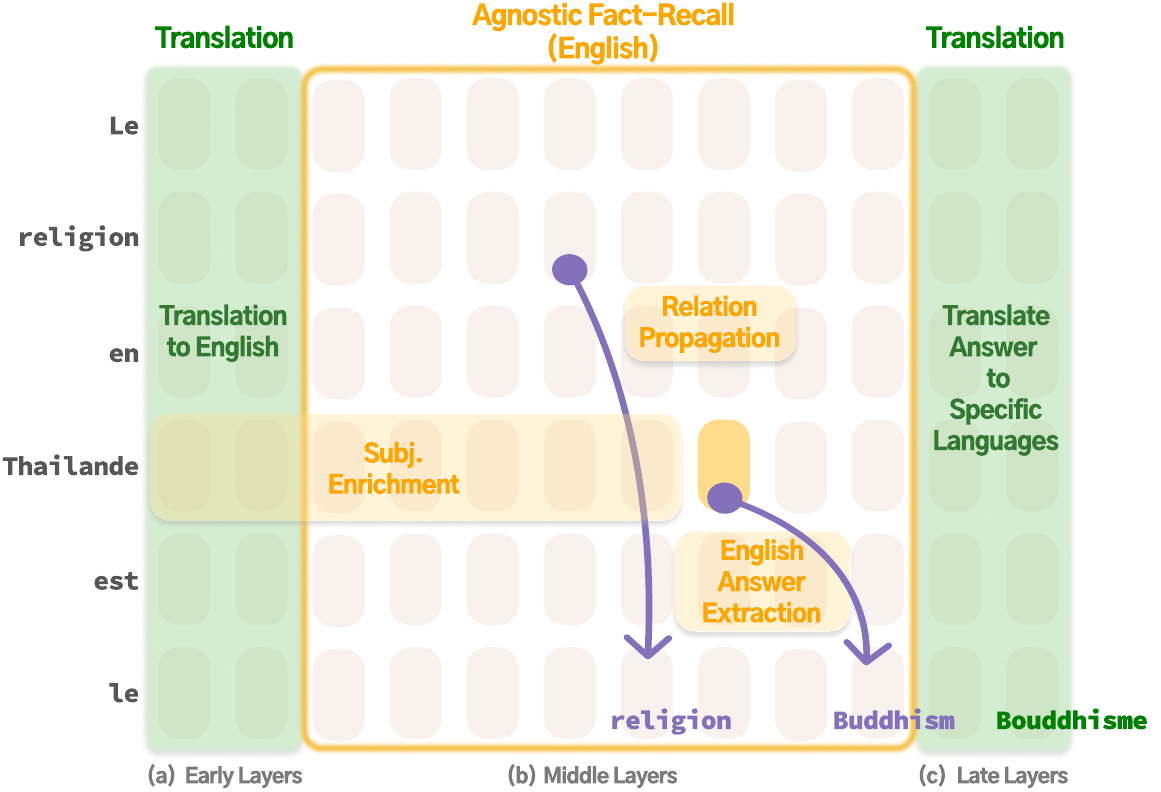}
    \caption{Hypothesized pipeline for multilingual factual recall. In this work, we focus on (1) the late-layer conversion highlighted in green on the right (\S\ref{sec:late-stage}) and (2) the English-centric factual-recall mechanism highlighted in yellow (adapted from~\citet{geva2023dissectingrecallfactualassociations} and see details in \S\ref{sec:early-stage}.)}
    \label{fig:pipeline}
\end{figure}

\begin{figure*}[t]
    \centering
    \includegraphics[width=\textwidth]{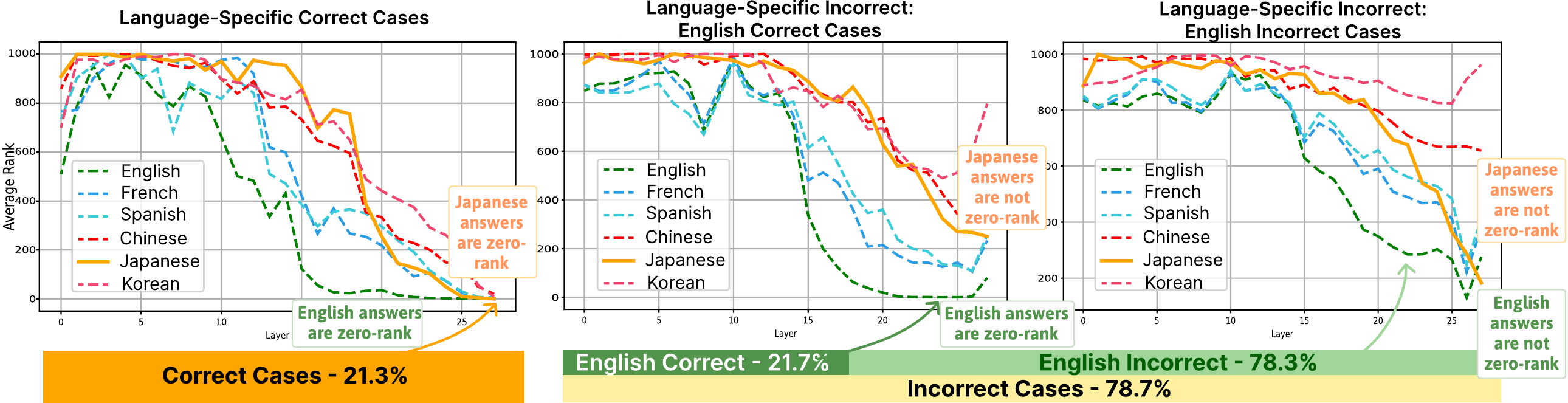}
    \caption{The bottom bar summarizes model performance on multilingual factual recall across languages. The figures above display average rank changes of answers by layer using Logit Lens with Japanese prompts. The left shows rank changes across correct instances. The middle and right show incorrect ones, which can be broken down to cases where intermediate English answers are right or wrong respectively.}
    \label{fig:pipeline_correct}
\end{figure*}

\paragraph{Factual Recall Datasets} Similar to previous works~\citep{geva2023dissectingrecallfactualassociations, constanza_multilingual, wang2025lost}, we represent each fact as a \texttt{(subject, relation, answer)} triple. The subject and relation are embedded in a natural language prompt, which is taken by the model as input; the model is expected to generate the answer as the next token. For example, the input for the fact triplet \texttt{(Thailand, main religion, Buddhism)} is \texttt{``The main religion in Thailand is''}, where the model should predict \texttt{``Buddhism''}. To study the recall mechanism at scale, we curate a factual dataset containing 2,862 validated\footnote{Each triple is manually validated to ensure it is correct.} triples that represent parallel facts across six languages (English, Chinese, Japanese, Korean, French, and Spanish). These languages are chosen to capture similarities and differences across diverse language families and writing scripts. Our dataset spans ten distinct relation types, including country languages, currencies, religions, and musicians' instruments, encompassing facts related to various geographical regions (See dataset details in Appendix~\ref{sec:appendix_dataset}).

\paragraph{Characterizing Multilingual Factual Recall} 
We use the logit lens~\citep{logit_lens} to understand how models arrive at the final answer during the factual recall process. Logit lens decodes the intermediate representations of an LM into tokens and has been widely used as a window to understand the internal processing pipeline~\citep{merullo2023language, wendler2024llamasworkenglishlatent, wu2024semantic, schut2025multilingual, ruochen_same_but_different, wang2025lost}. Specifically, we take the latent representation of each layer at the last token position and project it onto the vocabulary space by multiplying the unembedding matrix. Then, after applying the softmax function, we obtain the probability distribution for the next token prediction. This can be thought of as a ``print statement'' to see how the model is computing its final prediction across each intermediate layer of the forward pass. 

As previous works have pointed out~\citep{wendler2024llamasworkenglishlatent, zhao_handle_multilingualism, schut2025multilingual}, if the model primarily performs factual recall in English, we would expect the English answer to emerge as the top-ranked predicted token in the middle layers before the answer in the target language appears. In contrast, if the model operates in the target language or an interlingua, we would see either the target language as top-ranked throughout layers or no consistent pattern of language dominance. From Figure~\ref{fig:pipeline_correct} (left), when applying logit lens to Llama-3.2-3B~\citep{grattafiori2024llama} for the correct factual recall instances (21.3\% of all examples), we observe that the rank of the correct English answers (green) starts to decrease first around layer 10. At layer 21 in particular, the English answer is ranked as the top prediction on average, but from this point onward, the rank of the target language answer keeps decreasing and takes over at the very last layers. These observations suggest that, the model conducts factual knowledge retrieval in an ``English-centric'' concept space and only produces target language in the final decoding stages, supporting the hypothesized pipeline in Figure~\ref{fig:pipeline}.

But what happens in the remaining 78.7\% of cases where the model fails to produce the correct target-language answer? We further investigate the failure cases by applying the logit lens to the incorrect outputs. As shown in Figure~\ref{fig:pipeline_correct} (middle), for the first type, in 21.7\% of the error cases, the model successfully produces intermediate English answers around layer 21 but the target-specific answer never becomes a top-ranked prediction. The second case is the remaining 78.3\% instances (right), where the model is unable to retrieve the correct English answer therefore neither English or the target-language answer is top-ranked. We hypothesize that the first failure could result from insufficient late-stage translation where the second one is due to an underutilized English-centric recall mechanism.

We present this integrated hypothesized pipeline as intuition and motivation for subsequent work. In the next two sections, we further validate the pipeline by investigating potential causes of the failure points: does the model activate suboptimal components when processing non-English prompts, leading to translation and recall failures? We then propose targeted interventions to encourage correct translation (\S\ref{sec:late-stage}) and recall (\S\ref{sec:early-stage}) in order to mitigate these issues. See Section~\ref{sec:discussion} for further discussion of this pipeline and the questions it leaves open.

\section{Fixing Incorrect Translation Errors}
\label{sec:late-stage}

Above, we see that 21.7\% of errors appear to be due to bad translation--i.e., the model ``knows'' the answer in English yet outputs the wrong answer in the target language. In this section, we first investigate the pathway used to do internal language translation, and find that it is not the same as the pathway the model uses when prompted to translate directly (\S\ref{sec:translation-observation}). We then show that leveraging the model's translation pathway leads to significant performance increases (\S\ref{sec:translation-effect}).

\subsection{Translation Mechanism is Insufficiently Used}
\label{sec:translation-observation}

As shown in Section~\ref{sec:pipeline}, we notice that the model successfully produces intermediate English answers around layer 21, but fails to translate the answer back to the correct input-language answer.
To test whether the problem stems from an overall poor translation ability, we explicitly prompt the model to translate the expected answer into the target language directly.

We construct a parallel translation dataset aligned with the original factual recall examples. For each instance, we use the expected English answer to create a prompt for explicit translation (e.g. \texttt{Please translate this word into Spanish. Word: mammal, Translation:}), and expect the model's answer to be the same as the factual recall target answer (e.g. \texttt{mamífero}). The model can reach 56.1 accuracy on this translation task (See Appendix Figure~\ref{fig:appendix_fact_translation_accuracy} for more details) compared to 21.3 accuracy when being prompted for factual recall. This observation suggests that the model is capable of translating tokens to target languages accurately when being explicitly prompted, yet such capabilities are not fully leveraged in the factual recall context.

\begin{figure*}[t]
    \centering
    \includegraphics[width=1\linewidth]{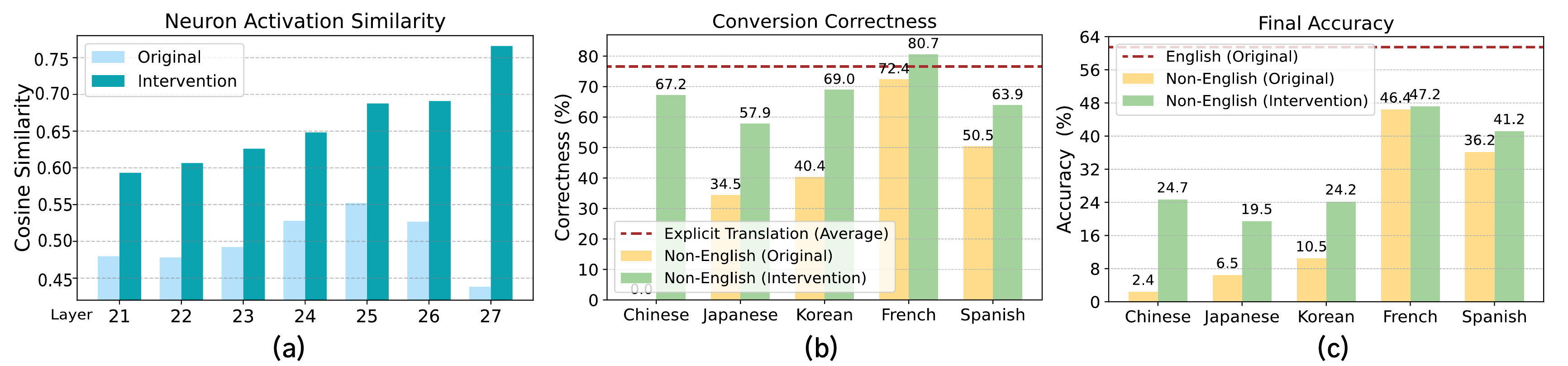}
        \caption{Effect of Translation Vector Intervention: (a) Neuron cosine similarity comparison between the recall task and translation task in late layers. (b) The rate comparison of correct final answers given correct intermediate English answers. (c) Recall task accuracy breakdown per language on the test set.}
    
    \label{fig:translate_three_results}
\end{figure*}

Motivated by this finding, we investigate whether there is a difference between the components used for explicit translation prompts and the components used to translate in the context of fact recall. (We henceforth refer to this latter translation process as \textit{conversion} when necessary to differentiate the two processes.) We conduct logit lens analysis and activation patching~\citep{NEURIPS2020_92650b2e} using TransformerLens~\citep{nanda2022transformerlens} on both tasks and observe a similar structural behavior in factual recall and explicit translation: the English answer token is shifted to the final position by around layer 17, and translating to input-language answer is predominantly handled by the MLP layers 22–27 (see Figure~\ref{fig:appendix_translation_logit_lens}, Appendix~\ref{sec:appendix_translation_fact_recall}). However, a closer inspection of neuron activation patterns reveals a crucial difference: although both tasks leverage layers 22–27, the cosine similarity between their MLP activations averages only 0.5 across layers (Figure~\ref{fig:translate_three_results} (a)). This indicates partial but not full overlap — the same layers are active, but the internal translation pathways differ.

Neuron similarity shows that the model is not engaging the most effective translation neurons unless explicitly prompted. However, can we extract a general signal to steer the model toward activating these components during factual recall? At layers 21–25, the last-token representation in both tasks encodes the same intermediate English answer, but only the explicit translation task moves toward a more accurate non-English representation. This suggests that, at the intermediate layer where English answers can be decoded, the representation carries not just shared semantic content but also a distinct task signal—one that activates different downstream pathways for fact-recall language conversion vs.\ explicit translation.

\subsection{Translation Difference Vector} 
Drawing from the difference-in-means concept editing approaches~\citep{diff_in_mean_belrose2023diffinmeans,refusal_direction}, we hypothesize that the difference between the model’s mean residual stream activations when processing fact-recall versus translation prompts can be used to nudge the model to activate a better translation route. Specifically, for each layer $\ell \in \mathcal{L}$ where $\mathcal{L} = \{21, 22, 23, 24, 25, 26,27\}$, we first compute the mean activation $\bar{h}_\mathcal{C}^{(\ell)}$ for all fact-recall prompts $p_\mathcal{C}\in\mathcal{C}$ and $\bar{h}_\mathcal{T}^{(\ell)}$ for all translation prompts $p_\mathcal{T}\in\mathcal{T}$ as follows\footnote{Note that both $\bar{h}_\mathcal{C}^{(\ell)}$ and $\bar{h}_\mathcal{T}^{(\ell)}$ are computed across relation datasets and languages. We also experiment with language-specific translation vectors, computed by averaging activations over translation prompts within each language, which yields comparable performance to the language-agnostic vector.}:

\begin{equation}
\bar{h}_\mathcal{C}^{(\ell)} = \frac{1}{|\mathcal{C}|} \sum_{p_\mathcal{C} \in \mathcal{C}} h_{p_{\mathcal{C}}}^{(\ell)} \quad\quad
\bar{h}_\mathcal{T}^{(\ell)} = \frac{1}{|\mathcal{T}|} \sum_{p_\mathcal{T} \in \mathcal{T}} h_{p_{\mathcal{T}}}^{(\ell)}
\end{equation}

\noindent We define the translation difference vector at layer $\ell$ as $\Delta^{(\ell)} = \bar{h}_\mathcal{T}^{(\ell)} - \bar{h}_\mathcal{C}^{(\ell)}$. To intervene, we add $\Delta^{(\ell)}$ to the residual stream of a multilingual fact-recall input at layer $\ell$ at the final token position\footnote{The intervention is computed using the residual stream value before layer processing and reinserted at the same point.}. 

\subsection{Effect of Translation Vector Intervention}
\label{sec:translation-effect}
To determine the most effective point of intervention, we evaluate translation difference vectors extracted from each of the layers 21 through 27. By testing each on a held-out validation set, we find that intervening at layer 21 yields the largest improvement in this stage's average performance (Appendix~\ref{sec:translation_vector_details}). We report both component- and task-level effects of this intervention on the test set.

As shown in Figure~\ref{fig:translate_three_results}(a), following the intervention, the cosine similarity between neuron activations during fact recall language conversion and those during explicit translation increases. This suggests on the component-level, the model’s internal behavior during multilingual fact-recall is being steered to more resemble that of the translation task. To evaluate performance improvement, we measure the conversion correctness rate, defined by the proportion of cases where the model correctly produces the final answer, conditioned on identifying the correct intermediate English answer. Figure~\ref{fig:translate_three_results}(b) shows that the intervention raises the average conversion correctness across all languages to an average of 67.74\%. Compared to the original 39.56\% conversion correctness, this intervention has significantly recovered the model’s translation capability across different languages.

This improvement at the conversion stage leads to a corresponding increase in factual recall accuracy across all languages, as shown in Figure~\ref{fig:translate_three_results}(c). We observe that the intervention has more significant impacts on language with non-Latin scripts (i.e. Chinese, Japanese and Korean) and modest impacts on French and Spanish, given their higher conversion rate prior to the intervention. 

The translation vector intervention supports the previous intuition about why translation failures occur: These failures are not due to a lack of translation capability; rather, when given a non-English prompt, the model lacks a signal to sufficiently integrate the optimal translation components in its fact-recall process. By injecting a single, general-purpose translation signal, we can recover much of the lost performance.

\section{Fixing Incorrect English Recall Errors}
\label{sec:early-stage}
Previously, we show that applying the translation difference vector intervention effectively corrects the conversion stage error illustrated in Figure~\ref{fig:pipeline_correct}(b). However, as shown in Figure~\ref{fig:pipeline_correct}(c), a different class of failure arises earlier: when given multilingual prompts, the model fails to retrieve the correct English answer, resulting in an incorrect final output even with better translation in late layers.

In this section, we investigate the middle recall stage (Figure~\ref {fig:pipeline}(b)) and identify that the English-centric factual recall pathway (Figure~\ref{fig:pipeline}) is underutilized in multilingual settings. We then introduce another vector-based intervention derived from in-context learning to improve the English-centric recall mechanism for multilingual factual recall.

\subsection{English Factual Recall Components are Insufficiently Activated}

\begin{figure*}[t]
    \centering
    \includegraphics[width=1\linewidth]{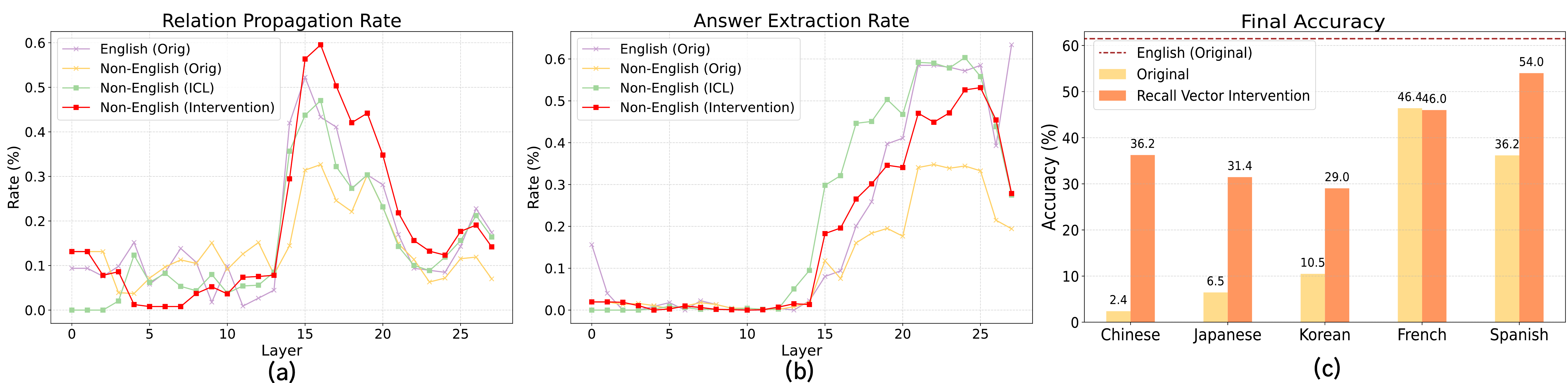}
    \caption{Effect of Recall Vector Intervention: (a) Intervention significantly improves the relation propagation substep (around layer 16). (b) English answer correctness: the intervention allows for more correct predictions of non-English final answers. (c) Task accuracy breakdown for all languages.}
    
    \label{fig:recall_three_results}
\end{figure*}

Based on prior work and our logit lens analysis, generating intermediate English answers suggests that the model might internally rely on a similar factual recall mechanism used for English prompts. We analyze the recall stage at a finer granularity: for both English and non-English prompts, we look at substages of the recall pipeline to understand whether non-English cases share the same mechanism and where inconsistencies may arise.
For English factual recall, one of the critical substages in early-middle layers is relation propagation~\citep {geva2023dissectingrecallfactualassociations}, which refers to the phenomenon that relation tokens get propagated to the final token position for final answer extraction. For example, as illustrated in Figure~\ref{fig:pipeline}(b), given the prompt ``\texttt{The official religion in Thailand is}'', the relation token ``\texttt{religion}'' should appear as the top-ranked decoded prediction at the last token position at intermediate layers. After the subsequent answer extraction event, the answer token ``\texttt{Buddhism}'' then replaces the relation token as the top-ranked prediction. To track this behavior, we further use the logit lens as a diagnostic tool to quantify the rate of relation propagation and answer extraction across layers.

First, for the relation propagation substage, we compute the rate of relation propagation as the proportion of examples in which English relation token (or equivalent\footnote{We account for cases where the relation expressed in the prompt spans multiple tokens and include synonymous forms. Implementation details are provided in \ref{sec:relation_toks} and \ref{sec:synonyms}.}) appears as the top-ranked prediction at the final token position. Comparing English and non-English prompts, we find that the relation information propagates at similar layers but English prompts have significantly higher rates. As shown in Figure~\ref{fig:recall_three_results}(a), at layer 16, relation tokens reach the final position in 43.30\% of English prompts (purple line), compared to only 32.65\% for multilingual prompts (yellow line). This discrepancy suggests that multilingual prompts trigger the relation propagation as English prompts, yet not as sufficiently.

Subsequently, we measure the answer extraction rate, defined as the percentage of instances across layers where the model’s top-ranked decoded token transitions to the correct English answer, indicating successful answer extraction.
In Fig.~\ref{fig:recall_three_results}(b), we see a consistent increase beginning at layer 15 and peaking at layer 21 for both English and non-English prompts. However, there exists a substantial gap where English prompts achieve significantly higher rates compared to non-English ones. 

While English factual recall prompts can more successfully activate the internal factual recall mechanism, propagating the relation token and then querying for extraction, the model sometimes fails to follow this path effectively when given non-English prompts. This comparison between English and non-English suggests that the recall inconsistencies result from earlier task recognition and relation token identification stages.

\subsection{Recall Task Vector}
Where and how can we provide a stronger signal to sufficiently activate the English-centric recall substages for multilingual cases? Prior work has highlighted the importance of in-context learning (ICL) for task recognition. In particular, \citet{sia2024doesincontexttranslationhappen} argues that in-context learning helps with identifying the task rather than learning it. Consistent with this view, in Figure~\ref{fig:recall_three_results}(a) and (b), we observe that given five-shot non-English ICL examples, both relation propagation and answer extraction significantly improve and become more aligned to the English rates (green line). 

Beyond explicit ICL examples, function vectors~\citep{todd2024functionvectorslargelanguage} and task vectors~\citep{task_vector} extracted from ICL can be injected into zero-shot runs to achieve comparable performance to ICL. Can we similarly derive a signal that helps to better activate intermediate English recall mechanisms under multilingual settings?

We construct a recall vector that aims to capture a general activation signal associated with English factual recall. Using all training instances with 5-shot ICL examples, we compute the average hidden activation $\bar{h}^\ell$ at the final token position, extracted at a specific layer $\ell$. We inject this averaged vector into the model’s residual stream at layer $\ell$ and then evaluate the test set. Different from prior work, this vector is extracted and applied for samples across different tasks (different relation-datasets and languages), so the vector is a task-independent signal that motivates a general recall behavior.\footnote{We also extract vectors specific to each dataset and test their effect. We find that dataset-specific vectors have comparable performance with the independent ones.}

\subsection{Effect of Recall Vectors}

To identify the optimal recall vector, we compute a set of candidate intervention vectors for each intervention layer $\ell \in [L]$ and scaling factor $i$\footnote{We experiment with scaling factors ranging from 1 to 5 because higher values introduce excessive noise and reduce answer quality.} (which controls the intensity of the injected signal). This results in $|I| \times L$ candidate vectors. Each candidate is evaluated on the validation set to assess its effectiveness in improving the model’s factual recall, specifically by measuring gains in intermediate English answer accuracy and relation propagation. The most effective vector is selected based on its ability to increase English answer correctness. As shown in Appendix~\ref{sec:recall_vector_details}, the optimal configuration corresponds to layer $\ell = 3$ with a scaling factor of $i=2$.

Using the best configuration, we observe that this dataset-independent recall vector triggers more relation propagation than ICL examples (Figure~\ref{fig:recall_three_results}(a) red line), resulting in a significant boost of successful extraction in (Figure~\ref{fig:recall_three_results}(a), red line). At the component level, we observe that after the intervention, the attention heads most important for English fact recall become more active during multilingual factual recall processing (Appendix Figure~\ref{fig:head_ablation},\ref{fig:top5heads_breakdown}). Furthermore, when decoding from the output vectors of attention modules, more correct English answers are directly outputted (Appendix Figure~\ref{fig:experiment_extraction}). These component-level change support our hypothesis that previous multilingual failures stem directly from insufficient engagement of these English factual recall components, and the general recall vector intervention delivers an effective signal in re-engaging the internal processing of the English factual recall mechanism, leading to better overall answer retrieval across all languages (Figure~\ref{fig:recall_three_results}(c)).

It's surprising that we are able to extract and apply a task-independent and language-independent recall vector to improve general zero-shot performance across ten diverse relations and five languages. This contrasts with previous studies on function and task vectors~\citep{todd2024functionvectorslargelanguage,task_vector}, which focus on vectors tailored to specific tasks (e.g., retrieving a country’s capital). Understanding how to extract such a generalizable signal requires further investigation, as it offers new insights into the granularity of the information these vectors encode.

\section{Intervention Effects}
\begin{figure*}[t]
    \centering
    \includegraphics[width=1\linewidth]{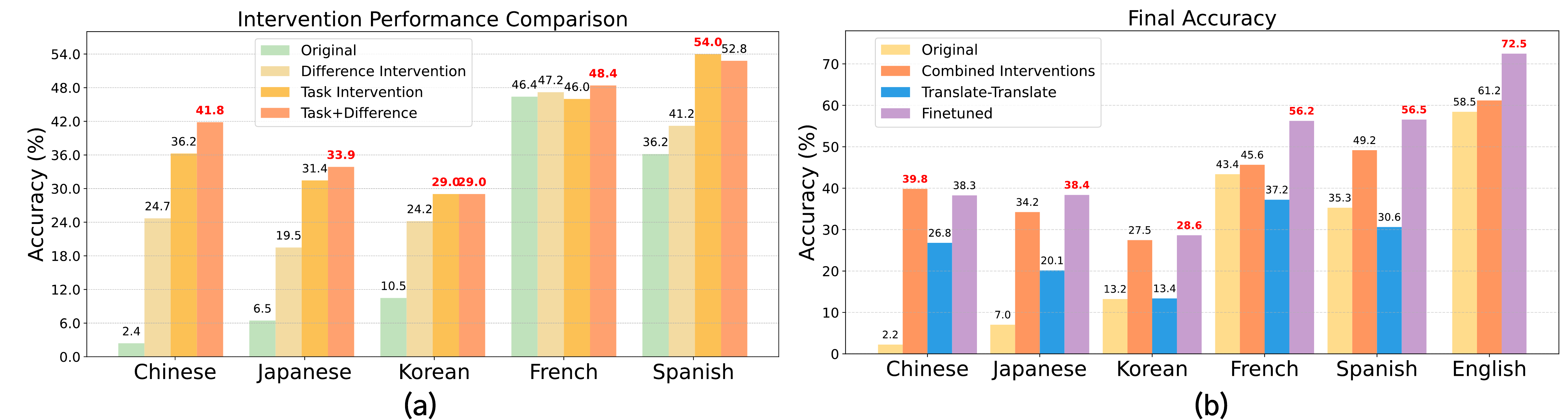}
        \caption{(a) Comparing the individual and combined effects of the translation and the recall vector. (b) Performance comparison between our intervention and baseline methods across three random data splits.}
    \label{fig:final_performance}
\end{figure*}

In Figure~\ref{fig:final_performance} (a), we compare the intervention effects of the translation vector, the recall vector, and their combination. For Chinese, Japanese, Korean, and French, we find that the combined intervention yields the highest final accuracy.\footnote{See Appendix~\ref{sec:appendix_intervention_comparison} for a detailed configuration and substage-level comparison of these effects.} 

We additionally compare our intervention to two non-mechanistic baselines. First, ``translate-recall-translate'' is a multi-step prompting strategy in which we query the model with three prompts sequentially: explicitly instructs the model to translate the question into English,
conduct the task in English, and then translate the response back to the target language~\citep{baseline_translate_huang,baseline_translate_shi}. For each example, we pass intermediate outputs from one step to the next and use the final output for evaluation. Second, we compare against fine-tuning, where we fine-tune the model on all training sets across languages and relations for 30 epochs and report the performance on the best checkpoint.\footnote{We also evaluate the generalization capabilities of our methods and baselines by holding out a subset of relations for testing. Further details can be found in the Appendix~\ref{sec:across_dataset_performance}.}

Our results in Figure~\ref{fig:final_performance} (b) demonstrate that our intervention consistently outperforms the translate-recall-translate baseline (detailed analysis in Appendix~\ref{sec:appendix_baseline_comparison}). While finetuning achieves higher overall accuracy, our approach remains competitive, particularly for languages in non-Latin scripts. This finding suggests that our training-free intervention improves upon prompting methods and, with performance comparable to finetuning, highlights its potential for robust cross-lingual knowledge retrieval without the need for additional training resources.

\section{Related Work}

As discussed in Section~\ref{sec:intro} and~\ref{sec:pipeline}, our work builds on prior investigations into factual recall mechanisms~\citep{geva2023dissectingrecallfactualassociations,meng2022locating,hase2023does,chughtai2024summing, yao2024knowledge} and multilingual processing in language models~\citep{conneau-etal-2020-emerging, muller-etal-2021-first, wendler2024llamasworkenglishlatent,wu2024semantic,schut2025multilingual,chughtai2024summing,constanza_multilingual,ruochen_same_but_different,ferrando2024similarity, wilie2025high}. Closely related concurrent work~\citep{wang2025lost} also addresses translation failures at the final generation stage. In contrast, our approach (1) introduces a language-agnostic intervention that generally activates more translation neurons instead of linear mapping between languages, and (2) additionally targets an earlier failure point in the factual recall pipeline, offering interventions at the intermediate "recall" stage that further validate the multi-step structure of multilingual factual retrieval.

Our intervention methods are inspired by previous works in steering vectors. They modulate model behavior at inference time by injecting learned vectors into intermediate activations~\citep{subramani2022extracting, turner2023steering, li2023inference, panickssery2023steering, sv_truthfulness,sv_sentiment, refusal_direction}, Other related vector-based intervention includes function and task vectors~\citep{todd2024functionvectorslargelanguage,task_vector},  which focus on understanding transformers as learning compact, concrete, and causal vector representations of higher-level functional concepts. Differently, we construct dataset-independent and language-independent vectors to strengthen latent pathways already present in the model. This goal aligns with component reuse approaches~\citep{olsson2022context, gurnee2023finding, merullo_component_reuse}, though we operate at a higher level—steering computation toward effective internal trajectories without explicitly localizing or reactivating individual components.

\section{Discussion}
\label{sec:discussion}
We describe a comprehensive pipeline which explains multilingual LLMs' factual recall mechanisms, integrating and extending findings from previous interpretability studies on both multilingual models and English factual recall. Using mechanistic insights from this pipeline, we identify sources of error and design targeted interventions. The predictable effects of these interventions support our hypothesis that multilingual LLMs process information through an English-centric concept space before generating language-specific responses. Our results raise several interesting directions warranting further investigation:

\paragraph{Understanding Early Layers} While our intervention improves the propagation of English relations in multilingual prompts, the precise connection between the language-specific translation stage and the subject enrichment substep in early layers remains unexplored. Our preliminary analysis reveals that non-relation tokens (approximately 13\% of subject tokens) also undergo translation to English in intermediate layers. However, the reliability of the logit lens for early-layer analysis is questionable~\citep{belrose2023eliciting, ghandeharioun2024patchscopes}. Future research would benefit from alternative analysis strategies that can more faithfully reflect model behavior in early layers.

\paragraph{In-context learning vs. Interventions} In Section~\ref{sec:early-stage}, our recall vectors extracted from ICL runs demonstrate positive improvement on multilingual factual recall tasks. However, standard 5-shot ICL outperforms our intervention-based method. This is expected, as ICL encodes more direct language and task-specific information compared to our language and task-agnostic vectors. Nevertheless, this raises questions about the relative merits of mechanistic interventions like we propose vs.\ more familiar ``black box'' techniques for influencing model behavior. Of course, providing multiple in-context examples can often be impractical in real-world applications--asking users to provide many fact recall examples in order to look up one fact would be cumbersome. Nonetheless, as increasing progress is made on mechanistic approaches, more work will be needed to determine the best method for achieving the desired end-system behavior.

\section{Conclusion}
We introduce targeted vector-based interventions that effectively reduce cross-lingual factual recall inconsistencies, validating the multilingual LLM processing pipeline observed in previous research. Our work represents an initial step toward leveraging mechanistic insights to direct the model toward better internal paths to unlock its hidden potential. Future research should look into developing automated methods to identify such weaknesses and implement corresponding solutions, improving the robustness of multilingual LLMs across diverse linguistic contexts.

\section*{Limitations}

Our study's scope is limited to five non-English languages and ten relations on a single model. We observe consistent performance gains across languages, with more significant improvements seen in languages with non-Latin scripts such as  Chinese, Japanese, and Korean. Future research should expand this investigation to more diverse language families and syntactic structures in order to determine how general the observed mechanism and interventions are. Additionally, quantifying how our findings apply to relations of varying complexity or different models would provide a more comprehensive understanding of multilingual factual recall mechanisms.

\section*{Ethical Considerations}
This study investigates the mechanisms behind multilingual factual recall in LLMs and proposes targeted interventions to address cross-lingual factual inconsistencies. The dataset used in our experiments is manually curated and thoroughly reviewed to ensure that it does not contain any personally identifiable information or sensitive data. Moreover, our proposed intervention methods can provide actionable insights on how to improve fairness and reduce bias across languages in existing multilingual LLMs. Future research is necessary to understand the generalizability and robustness of these methods through more comprehensive evaluation across additional languages, tasks, and model architectures.

\section*{Acknowledgment}
We thank Tianze Hua, Apoorv Khandelwal, Jack Merullo, Zhuonan Yang, Qinan Yu, Zheng-Xin Yong, Catherine Chen, William Rudman, Samuel Musker, Reza Esfandiarpoor and other Brown Superlab members for their helpful discussion and feedback on our work.

\bibliography{custom}

\newpage
\appendix
\label{sec:appendix}

\section{Dataset Construction}
\label{sec:appendix_dataset}

To study the factual recall mechanisms of language models at scale, we curate a multilingual dataset spanning ten different relations and five languages, as detailed below.

For each relation, we use o1\footnote{\url{https://openai.com/o1/}} to generate a list of 50 English \texttt{(subject, attribute)} pairs. We manually verify each generated triplet using Google Search to eliminate incorrect or ambiguous cases. We then prompt o1 to produce semantically equivalent prompts in six languages: English, Chinese, Japanese, Korean, French, and Spanish. All translations are manually verified to ensure accuracy. Because we focus on factual knowledge that holds across languages and cultures, the underlying entity in each triplet remains constant; only its surface linguistic form differs.

We explicitly instruct o1 to include facts associated with diverse geographical regions by prompting it to ``make sure to include factual pairs from all geographic regions''. This design choice allows us to test whether language models generalize factual knowledge across multilingual prompts, even when the facts are not centered around English-speaking regions. For instance, under the musician instrument relation, our dataset includes prompts about Asian musicians such as ``The primary instrument played by Joe Hisaishi is the''.

In total, our dataset comprises 2,862 verified fact triplets across six languages. The statistics per relation are shown in Table~\ref{tab:relation_distribution} and see prompt templates in Table~\ref{tab:prompt_example}. We will release our data upon publication under the CC-BY-SA 4.0 license.

\begin{table}[h]
\centering
\resizebox{1\columnwidth}{!}{%
\begin{tabular}{lc}
\hline
\textbf{Relation Type} & \textbf{Triplets per Language} \\
\hline
Country currency       & 51 \\
Country language       & 45 \\
Book language          & 54 \\
Animal classification  & 47 \\
Object color           & 43 \\
Country religion       & 46 \\
Language family        & 50 \\
Musician country       & 47 \\
Musician instruments   & 45 \\
Person university      & 49 \\
\hline
\end{tabular}
}
\caption{Distribution of relation types in our multilingual factual dataset.}
\label{tab:relation_distribution}
\end{table}

\definecolor{lightgray}{gray}{0.95}

\begin{table*}[h!]
\centering
\small
\rowcolors{2}{white}{lightgray}
\begin{tabularx}{\textwidth}{lX}
\toprule
\textbf{Relation Dataset} & \textbf{Multilingual Prompt Examples} \\
\midrule

\textbf{country\_language} & 
\begin{CJK*}{UTF8}{gbsn}加拿大的官方语言是\end{CJK*} \newline
The official language in Canada is \newline
La langue officielle au Canada est l' \newline
\begin{CJK*}{UTF8}{min}カナダの公用語は\end{CJK*} \newline
\begin{CJK*}{UTF8}{mj}캐나다의 공용어는\end{CJK*} \newline
La gente en Canadá habla \\
\midrule

\textbf{country\_religion} & 
\begin{CJK*}{UTF8}{gbsn}美国的主要宗教是\end{CJK*} \newline
The main religion practiced in United States is \newline
La principale religion pratiquée aux États-Unis est le \newline
\begin{CJK*}{UTF8}{min}アメリカ合衆国で主に信仰されている宗教は\end{CJK*} \newline
\begin{CJK*}{UTF8}{mj}미국에서 주로 믿는 종교는\end{CJK*} \newline
La religión principal practicada en Estados Unidos es el \\
\midrule

\textbf{language\_family} & 
\begin{CJK*}{UTF8}{gbsn}英语所属的语系是\end{CJK*} \newline
The language English belongs to the language family of \newline
Anglais appartient à la famille de langues \newline
\begin{CJK*}{UTF8}{min}英語が属する語族は\end{CJK*} \newline
\begin{CJK*}{UTF8}{mj}영어가 속하는 어족은\end{CJK*} \newline
Inglés pertenece a la familia de lenguas \\
\midrule

\textbf{musician\_country} & 
\begin{CJK*}{UTF8}{gbsn}路德维希·范·贝多芬出生的国家名为\end{CJK*} \newline
The birth country of Ludwig van Beethoven is \newline
Le pays de naissance de Ludwig van Beethoven est l' \newline
\begin{CJK*}{UTF8}{min}ルートヴィヒ・ヴァン・ベートーヴェンの出身国は\end{CJK*} \newline
\begin{CJK*}{UTF8}{mj}루트비히 반 베토벤의 출생 국가는\end{CJK*} \newline
El país de nacimiento de Ludwig van Beethoven es \\
\midrule

\textbf{musician\_instruments} & 
\begin{CJK*}{UTF8}{gbsn}路德维希·范·贝多芬主要演奏的乐器名叫\end{CJK*} \newline
The primary instrument played by Ludwig van Beethoven is the \newline
L'instrument principal joué par Ludwig van Beethoven est le \newline
\begin{CJK*}{UTF8}{min}ルートヴィヒ・ヴァン・ベートーヴェンが主に演奏する楽器は\end{CJK*} \newline
\begin{CJK*}{UTF8}{mj}루트비히 반 베토벤가 주로 연주하는 악기는\end{CJK*} \newline
El instrumento principal que toca Ludwig van Beethoven es el \\
\midrule

\textbf{object\_color} & 
\begin{CJK*}{UTF8}{gbsn}香蕉的颜色是\end{CJK*} \newline
Banana has a color of \newline
La couleur de Banane est \newline
\begin{CJK*}{UTF8}{min}バナナの色は\end{CJK*} \newline
\begin{CJK*}{UTF8}{mj}바나나의 색깔은\end{CJK*} \newline
El color de Banana es \\
\midrule

\textbf{person\_university} & 
\begin{CJK*}{UTF8}{gbsn}村上春树就读的大学名叫\end{CJK*} \newline
The college that Haruki Murakami attended was called \newline
L'université où Haruki Murakami a étudié s'appelle \newline
\begin{CJK*}{UTF8}{min}村上春樹が通った大学の名前は\end{CJK*} \newline
\begin{CJK*}{UTF8}{mj}무라카미 하루키이 다녔던 대학의 이름은\end{CJK*} \newline
La universidad a la que asistió Haruki Murakami se llama \\
\midrule

\textbf{country\_currency} & 
\begin{CJK*}{UTF8}{gbsn}巴西的官方货币是\end{CJK*} \newline
The official currency of Brazil is called the \newline
La monnaie officielle de Brasil s'appelle \newline
\begin{CJK*}{UTF8}{min}ブラジルの公式通貨は\end{CJK*} \newline
\begin{CJK*}{UTF8}{mj}브라질의 공식 화폐는\end{CJK*} \newline
La moneda oficial de Brésil se llama \\
\midrule

\textbf{book\_language} & 
\begin{CJK*}{UTF8}{gbsn}伊利亚特最初编写时使用的语言为\end{CJK*} \newline
The language that The Iliad was originally written in was \newline
La langue dans laquelle L’Iliade a été écrit à l'origine était le\newline
\begin{CJK*}{UTF8}{min}イーリアスが最初に書かれた言語は\end{CJK*} \newline
\begin{CJK*}{UTF8}{mj}그리스어가 원래 작성된 언어는\end{CJK*} \newline
El idioma en el que griego fue escrito originalmente es \\
\midrule

\textbf{animal\_classification} & 
\begin{CJK*}{UTF8}{gbsn}大象在生物学上被分类为一种\end{CJK*} \newline
Elephant is biologically classified as a \newline
Éléphant est biologiquement classé comme un \newline
\begin{CJK*}{UTF8}{min}象は生物学的に分類される\end{CJK*} \newline
\begin{CJK*}{UTF8}{mj}코끼리는 생물학적으로 분류된다\end{CJK*} \newline
Elefante está clasificado biológicamente como un \\
\bottomrule
\end{tabularx}
\caption{Examples of multilingual prompts for each dataset.}
\label{tab:prompt_example}
\end{table*}

\subsection{Relation tokens for each dataset}
The relation in the prompt may span multiple tokens, so we manually select the most informative tokens for each relation, as specified in Table~\ref{tab:relation_words_summary}.
\label{sec:relation_toks}
\definecolor{lightgray}{gray}{0.95}

\begin{table*}[h!]
\centering
\small
\rowcolors{2}{white}{lightgray}
\begin{tabularx}{\textwidth}{lX}
\toprule
\textbf{Relation Dataset} & \textbf{Multilingual Relation Words} \\
\midrule

\textbf{person\_university} & 
college, attended / \begin{CJK*}{UTF8}{gbsn}大学, 就读\end{CJK*} / \begin{CJK*}{UTF8}{min}大学, 通った\end{CJK*} / \begin{CJK*}{UTF8}{mj}대학, 다녔던\end{CJK*} / universidad, asisti\'o / universit\'e, \'{e}tudi\'e \\
\midrule

\textbf{country\_currency} & 
currency / \begin{CJK*}{UTF8}{gbsn}货币\end{CJK*} / \begin{CJK*}{UTF8}{min}通貨\end{CJK*} / \begin{CJK*}{UTF8}{mj}화폐\end{CJK*} / moneda / monnaie \\
\midrule

\textbf{book\_language} & 
language, written, original / \begin{CJK*}{UTF8}{gbsn}语言, 编写\end{CJK*} / \begin{CJK*}{UTF8}{min}言語, 書かれた\end{CJK*} / \begin{CJK*}{UTF8}{mj}언어, 작성된\end{CJK*} / idioma, escrito / langue, \'{e}crit \\
\midrule

\textbf{animal\_classification} & 
classified, biologically / \begin{CJK*}{UTF8}{gbsn}分类, 生物学\end{CJK*} / \begin{CJK*}{UTF8}{min}分類, 生物学的\end{CJK*} / \begin{CJK*}{UTF8}{mj}분류, 생물학적으로\end{CJK*} / clasificado, biol\'ogicamente / class\'e, biologiquement \\
\midrule

\textbf{country\_language} & 
language / \begin{CJK*}{UTF8}{gbsn}语言\end{CJK*} / \begin{CJK*}{UTF8}{min}公用語\end{CJK*} / \begin{CJK*}{UTF8}{mj}공용어\end{CJK*} / idioma / langue \\
\midrule

\textbf{country\_religion} & 
religion, practiced / \begin{CJK*}{UTF8}{gbsn}宗教\end{CJK*} / \begin{CJK*}{UTF8}{min}宗教, 信仰\end{CJK*} / \begin{CJK*}{UTF8}{mj}종교, 믿는\end{CJK*} / religi\'on, practicada / religion, pratiqu\'ee \\
\midrule

\textbf{language\_family} & 
language, family / \begin{CJK*}{UTF8}{gbsn}语系\end{CJK*} / \begin{CJK*}{UTF8}{min}語族\end{CJK*} / \begin{CJK*}{UTF8}{mj}어족\end{CJK*} / lenguas, familia / langues, famille \\
\midrule

\textbf{musician\_country} & 
birth, country / \begin{CJK*}{UTF8}{gbsn}出生, 国家\end{CJK*} / \begin{CJK*}{UTF8}{min}出身, 国\end{CJK*} / \begin{CJK*}{UTF8}{mj}출생, 국가\end{CJK*} / nacimiento, pa\'{i}s / naissance, pays \\
\midrule

\textbf{musician\_instruments} & 
instrument, played / \begin{CJK*}{UTF8}{gbsn}乐器, 演奏\end{CJK*} / \begin{CJK*}{UTF8}{min}楽器, 演奏\end{CJK*} / \begin{CJK*}{UTF8}{mj}악기, 연주\end{CJK*} / instrumento, toca / instrument, jou\'e \\
\midrule

\textbf{object\_color} & 
color / \begin{CJK*}{UTF8}{gbsn}颜色\end{CJK*} / \begin{CJK*}{UTF8}{min}色\end{CJK*} / \begin{CJK*}{UTF8}{mj}색깔\end{CJK*} / color / couleur \\
\bottomrule
\end{tabularx}
\caption{Relation words for each fact recall dataset.}
\label{tab:relation_words_summary}
\end{table*}

\subsection{Selecting Relation Token Equivalents}
\label{sec:synonyms}
\begin{table*}[t]
\centering

\resizebox{1\linewidth}{!}{%
\begin{tabularx}{\textwidth}{p{0.95\textwidth}}
\toprule
\textbf{1.0} — Exact match with the reference word \\
\textbf{0.8–0.99} — Conceptual synonym or close paraphrase (e.g., ``hue'' for ``color'', ``dialect'' for ``language'') \\
\textbf{0.5–0.8} — Loosely related or contextually associated term (e.g., ``paint'' for ``color'', ``accent'' for ``language'') \\
\textbf{< 0.5} — Category member or specific instance of the concept (e.g., ``red'' for ``color'', ``yen'' for ``currency'', ``Spanish'' for ``language'') \\
\textbf{< 0.2} — Unrelated or irrelevant term \\
\textit{Note:} If a token appears to be a truncated or partial form of a meaningful word (e.g., ``pigm'' for ``pigment'', ``forgot'' for ``forget''), we infer that it is likely a lemmatized form and score based on its intended meaning. \\
\bottomrule
\end{tabularx}
}
\caption{GPT-4o Scoring Guidelines}
\label{tab:relation_synonym_rubric_tab}
\end{table*}

Since relation tokens are often not proper nouns, there can be multiple valid translations for a given non-English relation token. To address this, we employ a two-stage filtering and scoring process. First, we use spaCy~\citep{spacy} to lemmatize the predicted token. Then, we use WordNet~\citep{wordnet} to compute semantic similarity between the lemmatized predicted token and the English relation token. Tokens with zero similarity (i.e., clearly irrelevant) are discarded. For the remaining candidates, we prompt GPT-4o to assign a similarity score following a procedure inspired by~\citet{schut2025multilingual}.

We instruct GPT-4o to rate each candidate word from 0 to 1 based on its semantic similarity to the reference concepts, using the following rubrics in Table~\ref{tab:relation_synonym_rubric_tab}. We consider a token to be an acceptable equivalent of the English relation word if its GPT-4o-assigned score exceeds 0.8.

\clearpage
\section{Logit Lens}

\begin{figure*}[h]
    \centering
    \includegraphics[width=1.0\linewidth]{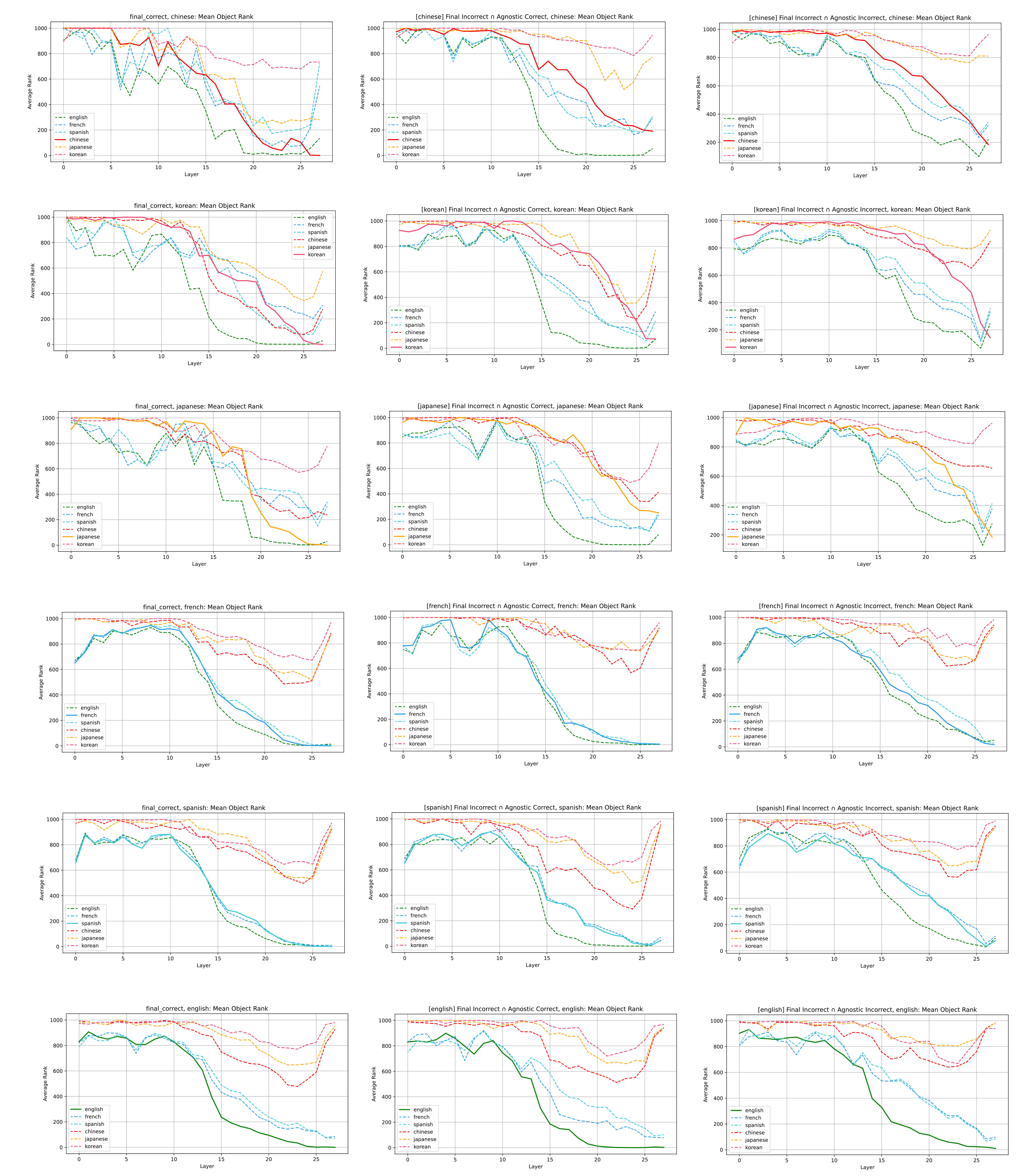}
    \caption{Language Breakdown of Answer Rank Changes Across Layers.}
    \label{fig:appendix_fact_recall_logit_lens_datasets}
\end{figure*}

\subsection{Detailed Breakdown of Model Failure Cases}

In Section~\ref{sec:pipeline}, we analyze failure cases by checking whether the model correctly identifies the intermediate English answer at layer 21. Due to tokenization, we consider the predicted token correct if it either appears within the correct answer string or if the correct answer is contained within the predicted token. If the model’s top-1 prediction at that layer matches the correct English answer, we categorize it as \textit{agnostic correct}. 

Importantly, this form of intermediate correctness can be examined not only at layer 21 but also across layers 20 to 27. That is, at each of these layers, we can assess whether the model internally “knows” the correct English answer, regardless of the final output. Tables~\ref{tab:pipeline_20} through~\ref{tab:pipeline_27} provide a detailed breakdown of model failures and agnostic correctness across these layers.

\begin{table}[h]
\centering

\label{tab:agnostic_final_breakdown_layer20}
\resizebox{1\columnwidth}{!}{%
\begin{tabular}{l r}
\toprule
\textbf{Category} & \textbf{Count (\%)} \\
\midrule
Total evaluated & 2385 \\
Agnostic correct & 504 (21.13\%) \\
Agnostic incorrect & 1881 (78.87\%) \\
\midrule
Final correct $\cap$ Agnostic correct & 279 (11.70\%) \\
Final incorrect $\cap$ Agnostic correct & 225 (9.43\%) \\
Final correct $\cap$ Agnostic incorrect & 229 (9.60\%) \\
Final incorrect $\cap$ Agnostic incorrect & 1652 (69.27\%) \\
\bottomrule
\end{tabular}
}
\caption{Global summary of agnostic correctness and final prediction correctness at Layer 20.}
\label{tab:pipeline_20}
\end{table}

\begin{table}[h]
\centering
\resizebox{1\columnwidth}{!}{%
\begin{tabular}{l r}
\toprule
\textbf{Category} & \textbf{Count (\%)} \\
\midrule
Total evaluated & 2385 \\
Agnostic correct & 792 (33.21\%) \\
Agnostic incorrect & 1593 (66.79\%) \\
\midrule
Final correct $\cap$ Agnostic correct & 385 (16.14\%) \\
Final incorrect $\cap$ Agnostic correct & 407 (17.06\%) \\
Final correct $\cap$ Agnostic incorrect & 123 (5.16\%) \\
Final incorrect $\cap$ Agnostic incorrect & 1470 (61.64\%) \\
\bottomrule
\end{tabular}
}

\caption{Global summary of agnostic correctness and final prediction correctness at Layer 21.}
\label{tab:agnostic_final_breakdown_percent}
\end{table}
\begin{table}[h]
\centering
\resizebox{1\columnwidth}{!}{%
\begin{tabular}{l r}
\toprule
\textbf{Category} & \textbf{Count (\%)} \\
\midrule
Total evaluated & 2385 \\
Agnostic correct & 867 (36.35\%) \\
Agnostic incorrect & 1518 (63.65\%) \\
\midrule
Final correct $\cap$ Agnostic correct & 407 (17.06\%) \\
Final incorrect $\cap$ Agnostic correct & 460 (19.29\%) \\
Final correct $\cap$ Agnostic incorrect & 101 (4.23\%) \\
Final incorrect $\cap$ Agnostic incorrect & 1417 (59.41\%) \\
\bottomrule
\end{tabular}
}
\caption{Global summary of agnostic correctness and final prediction correctness at Layer 22.}
\label{tab:agnostic_final_breakdown_layer22}
\end{table}
\begin{table}[h]
\centering
\resizebox{1\columnwidth}{!}{%
\begin{tabular}{l r}
\toprule
\textbf{Category} & \textbf{Count (\%)} \\
\midrule
Total evaluated & 2385 \\
Agnostic correct & 845 (35.43\%) \\
Agnostic incorrect & 1540 (64.57\%) \\
\midrule
Final correct $\cap$ Agnostic correct & 400 (16.77\%) \\
Final incorrect $\cap$ Agnostic correct & 445 (18.66\%) \\
Final correct $\cap$ Agnostic incorrect & 108 (4.53\%) \\
Final incorrect $\cap$ Agnostic incorrect & 1432 (60.04\%) \\
\bottomrule
\end{tabular}
}
\caption{Global summary of agnostic correctness and final prediction correctness at Layer 23.}
\label{tab:agnostic_final_breakdown_layer23}
\end{table}
\begin{table}[h]
\centering
\resizebox{1\columnwidth}{!}{%
\begin{tabular}{l r}
\toprule
\textbf{Category} & \textbf{Count (\%)} \\
\midrule
Total evaluated & 2385 \\
Agnostic correct & 823 (34.51\%) \\
Agnostic incorrect & 1562 (65.49\%) \\
\midrule
Final correct $\cap$ Agnostic correct & 387 (16.23\%) \\
Final incorrect $\cap$ Agnostic correct & 436 (18.28\%) \\
Final correct $\cap$ Agnostic incorrect & 121 (5.07\%) \\
Final incorrect $\cap$ Agnostic incorrect & 1441 (60.42\%) \\
\bottomrule
\end{tabular}
}
\caption{Global summary of agnostic correctness and final prediction correctness at Layer 24.}
\label{tab:agnostic_final_breakdown_layer24}
\end{table}
\begin{table}[h]
\centering
\resizebox{1\columnwidth}{!}{%
\begin{tabular}{l r}
\toprule
\textbf{Category} & \textbf{Count (\%)} \\
\midrule
Total evaluated & 2385 \\
Agnostic correct & 820 (34.38\%) \\
Agnostic incorrect & 1565 (65.62\%) \\
\midrule
Final correct $\cap$ Agnostic correct & 358 (15.01\%) \\
Final incorrect $\cap$ Agnostic correct & 462 (19.37\%) \\
Final correct $\cap$ Agnostic incorrect & 150 (6.29\%) \\
Final incorrect $\cap$ Agnostic incorrect & 1415 (59.33\%) \\
\bottomrule
\end{tabular}
}
\caption{Global summary of agnostic correctness and final prediction correctness at Layer 25.}
\label{tab:agnostic_final_breakdown_layer25}
\end{table}
\begin{table}[h]
\centering
\resizebox{1\columnwidth}{!}{%
\begin{tabular}{l r}
\toprule
\textbf{Category} & \textbf{Count (\%)} \\
\midrule
Total evaluated & 2385 \\
Agnostic correct & 1237 (51.87\%) \\
Agnostic incorrect & 1148 (48.13\%) \\
\midrule
Final correct $\cap$ Agnostic correct & 280 (11.74\%) \\
Final incorrect $\cap$ Agnostic correct & 957 (40.13\%) \\
Final correct $\cap$ Agnostic incorrect & 228 (9.56\%) \\
Final incorrect $\cap$ Agnostic incorrect & 920 (38.57\%) \\
\bottomrule
\end{tabular}
}
\caption{Global summary of agnostic correctness and final prediction correctness at Layer 26.}
\label{tab:agnostic_final_breakdown_layer26}
\end{table}

\begin{table}[h]
\centering
\label{tab:agnostic_final_breakdown_layer27}
\resizebox{1\columnwidth}{!}{%
\begin{tabular}{l r}
\toprule
\textbf{Category} & \textbf{Count (\%)} \\
\midrule
Total evaluated & 2385 \\
Agnostic correct & 1092 (45.79\%) \\
Agnostic incorrect & 1293 (54.21\%) \\
\midrule
Final correct $\cap$ Agnostic correct & 279 (11.70\%) \\
Final incorrect $\cap$ Agnostic correct & 813 (34.09\%) \\
Final correct $\cap$ Agnostic incorrect & 229 (9.60\%) \\
Final incorrect $\cap$ Agnostic incorrect & 1064 (44.61\%) \\
\bottomrule
\end{tabular}
}
\caption{Global summary of agnostic correctness and final prediction correctness at Layer 27.}
\label{tab:pipeline_27}
\end{table}

\section{Fixing Translation Error}

\begin{figure}[h]
    \centering
    \includegraphics[width=1.0\linewidth]{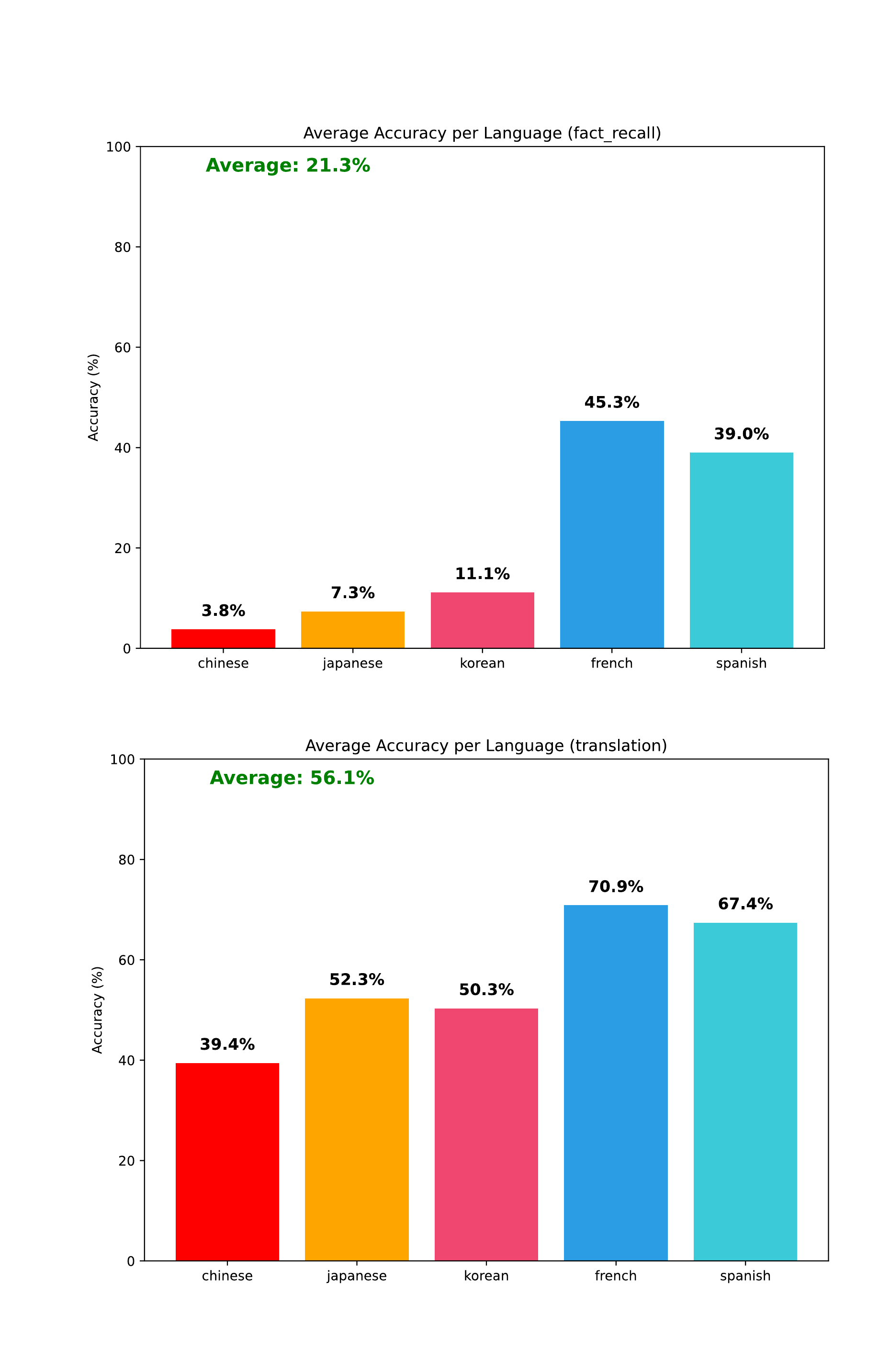}
    \caption{Fact Recall and Explicit Translation Performance Comparison on all data.}
    \label{fig:appendix_fact_translation_accuracy}
\end{figure}

\subsection{Explicit Translation Dataset Construction}
We adapt our fact-recall datasets into a translation task dataset. Specifically, we extract each [answer] from all fact-recall samples and format them into this prompt: ``Please translate this word into Chinese. Word:[answer], Translation:''. For example, for the answer ``mammal'' from the animal classification dataset, we create a translation variant prompt:  Please translate this word into Spanish. Word: `mammal', Translation:', and expect the correct answer to be ``mamífero''.

\subsection{Comparison between Translation and Conversion on Component-Level}
\label{sec:appendix_translation_fact_recall}
\begin{figure*}[h]
    \centering
    \includegraphics[width=1.0\linewidth]{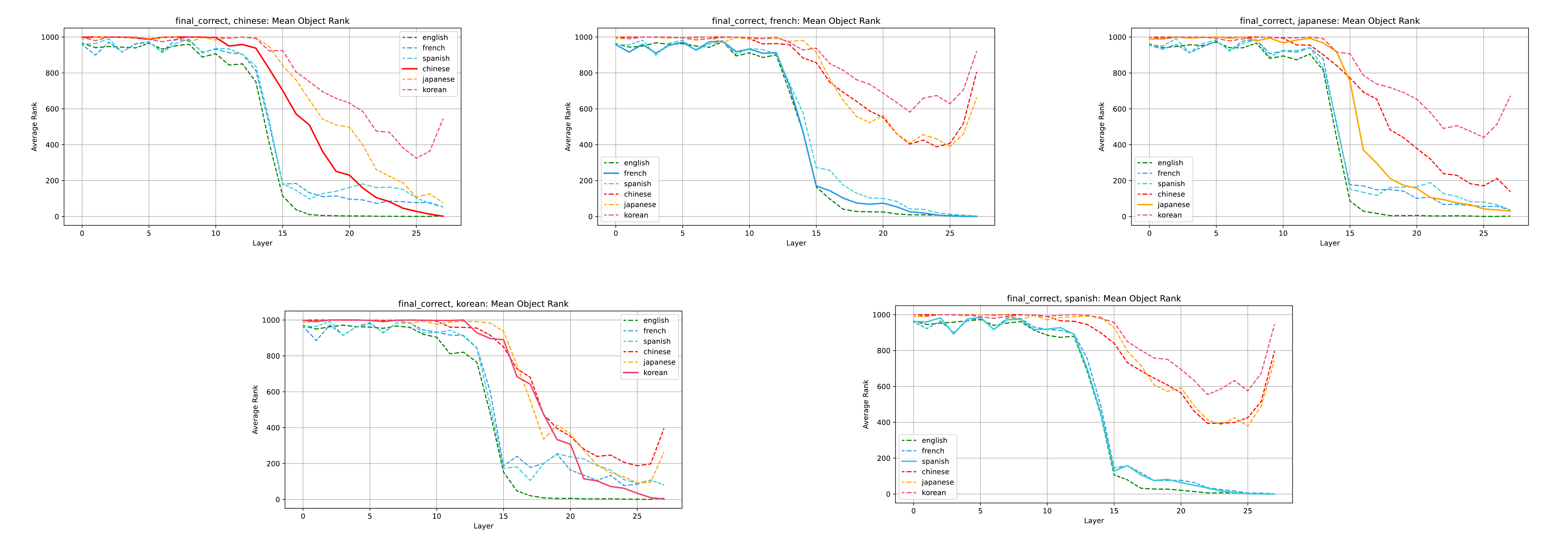}
    \caption{Logit lens on translation dataset reveals that the English answer has been moved from its original position to the last token position at around layer 15, and translation mechanism starts happening also at layer 15 when the translated answer slowly goes to zero-rank at the very end.}
    \label{fig:appendix_translation_logit_lens}
\end{figure*}

We apply logit lens analysis on our parallel translation dataset and observe \textit{overlapping} layer usage between translation and fact-recall conversion. Specifically, when the model is prompted to translate a word from English into another language, the English word is first shifted to the final token position by layer 17, and the translation process begins around layers happens from layer 17 and last until layer 25 (Appendix Figure ~\ref{fig:appendix_translation_logit_lens}). 

To further investigate whether the same model components are involved, we conduct activation patching~\citep{NEURIPS2020_92650b2e} using TransformerLens~\citep{nanda2022transformerlens}. We find that, in both translation and fact-recall conversion, MLP neurons in the later layers are most critical. Specifically, we apply the \textit{Activation Patching} framework to compute the \textit{Average Indirect Effect (AIE)} for each component. AIE quantifies the extent to which restoring a specific hidden state (e.g., an attention head or MLP block) from the clean input reduces the prediction error introduced by a corrupted input. Specifically, for a given output token $o$, AIE measures the fraction of the gap between the clean and corrupted predictions that is recovered by restoring only a single component. Formally:

\[
\mathrm{AIE} = \frac{P^{\ast, \text{clean } h^{(\ell)}_i}[o] - P^{\ast}[o]}{P[o] - P^{\ast}[o]}
\]

where $P[o]$ is the probability assigned to the correct output by the clean model, $P^{\ast}[o]$ is the probability under the corrupted input, and $P^{\ast, \text{clean } h^{(\ell)}_i}[o]$ is the probability when only component $h^{(\ell)}_i$ is restored to its clean state.

\begin{figure*}[h]
    \centering
    \includegraphics[width=1.0\linewidth]{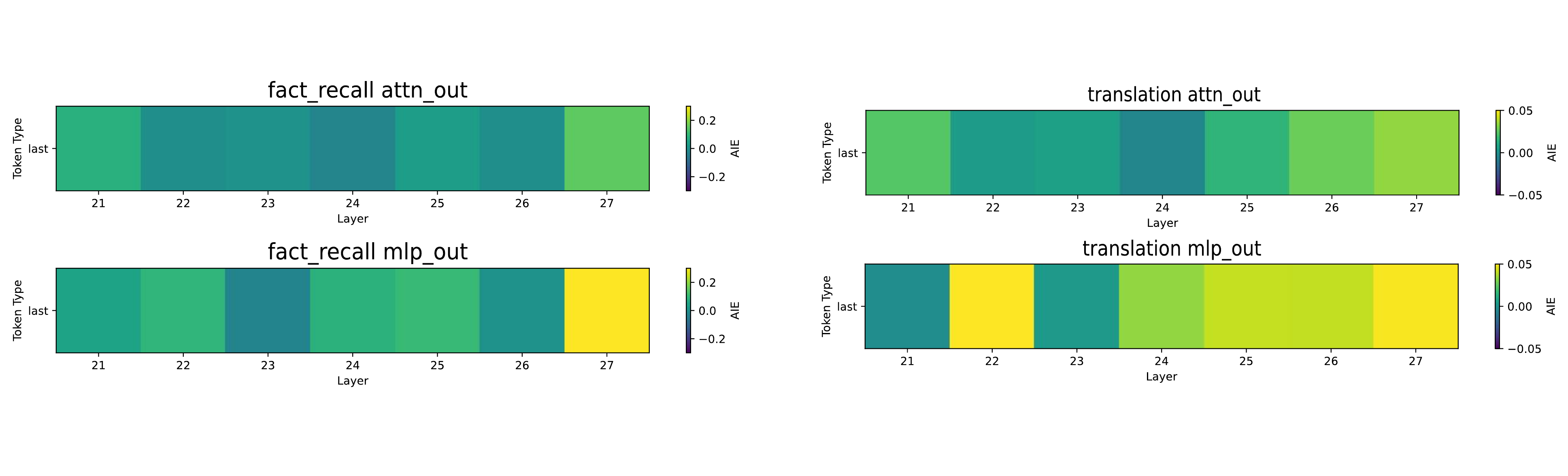}
    \caption{Average Indirect Effect of Patching Clean Component into Corrupted Runs. Left: running Activation Patching on fact-recall examples. Right: running Activation Patching on translation examples. From layer 21 to layer 27, MLP components exhibit more important effects.}
    \label{fig:appendix_translation_act_patching}
\end{figure*}

We compute AIE across all correct instances from both translation and fact-recall datasets, patching into each attention and MLP component across layers 22-25. As shown in Figure~\ref{fig:appendix_translation_act_patching}, we find that on MLP components have an average AIE of \textit{9.82\%}, while attention heads in the same position exhibit a much lower average AIE of \textit{2.74\%}. This indicates that the late-site MLP blocks contribute more to conversion and translation. These results highlight that while both translation and conversion tasks rely on similar regions and components of the model. This finding aligns with prior work showing that language-specific neurons on late-site MLPs~\citep{tang2024language, ruochen_same_but_different} and is also consistent with recent work by~\citet{constanza_multilingual}, which highlights the importance of late MLPs in the multilingual fact recall process.

\subsubsection{Hyperparameter for Translation Difference Vector}
\label{sec:hyperparam_translation}

\begin{figure}[h]
    \centering
    \includegraphics[width=1\linewidth]{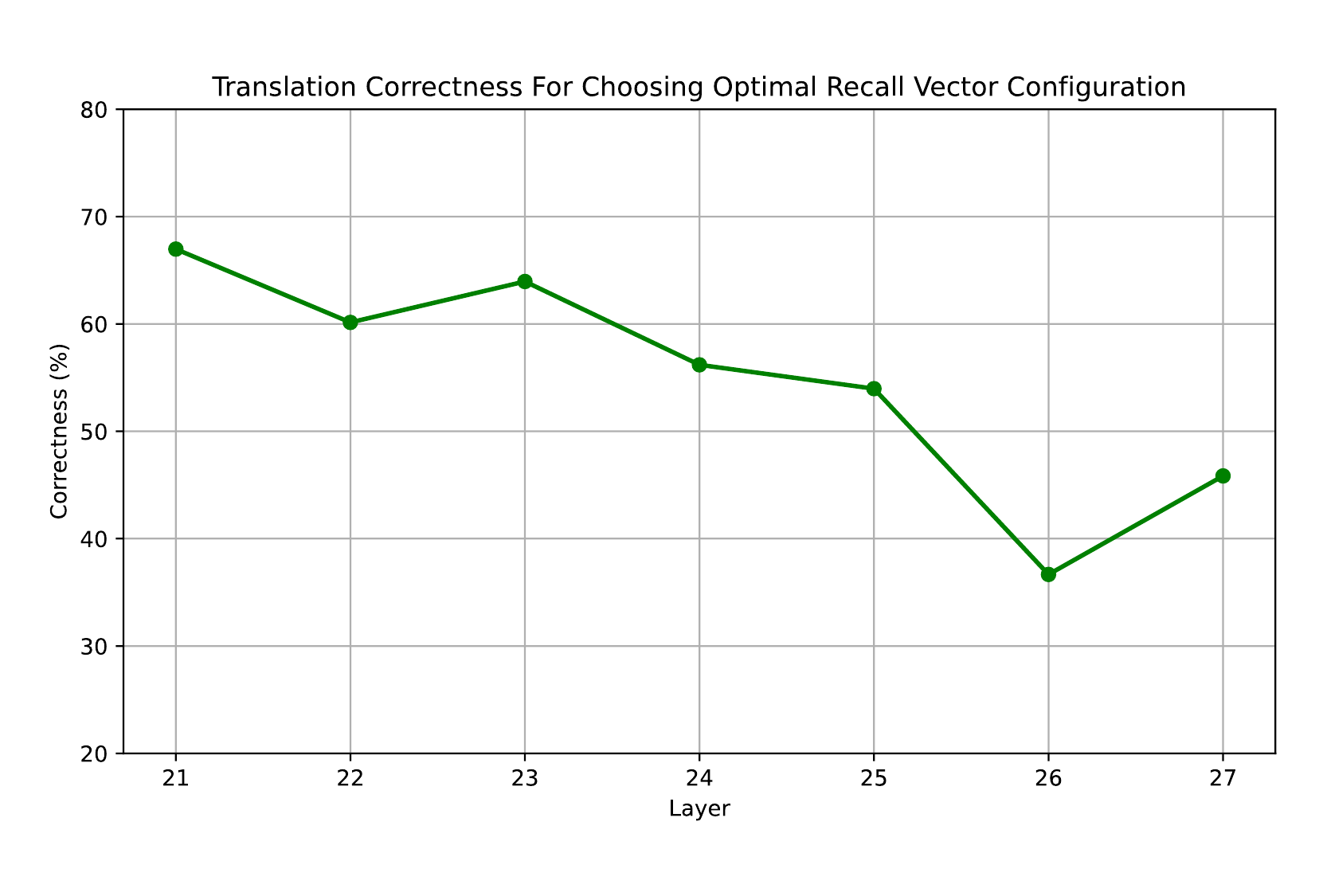}
    \caption{Translation Correctness when intervening at different layers.}
    
    \label{fig:translation_opt_hyperparam}
\end{figure}

Shown in Figure~\ref{fig:translation_opt_hyperparam}, when we extract and perform translation difference vector intervention on different layers, we test the translation correctness on the validation set and determine the best difference vector intervention layer is layer 21.

%translation_opt_hyperparam

\subsubsection{Translation Vector Effect Additional Details}
\label{sec:translation_vector_details}
In addition to gains in final correctness, as a side effect, we also observe an increase in agnostic correctness (Table~\ref{tab:tab_all_interventions}). Since this evaluation is conducted on a strictly held-out test set, these improvements are not due to data leakage. Through qualitative analysis, we find that in many cases where agnostic correctness improves, the model appears less confused at the final token. For example, prior to intervention, the top-1 token predicted via logit lens is often nonsensical (e.g., “WHAT”). After applying the intervention, the model instead generates a meaningful answer such as “mammals.” We hypothesize that the intervention reduces confusion in the later layers, allowing the model to project more confidently into the correct answer space. It is also possible that the intervention indirectly reinforces the model's tendency to map outputs through an internal English representation before translating into the target language, thereby enhancing its overall consistency.
\clearpage
\section{Fixing Recall Errors}
\begin{figure}[h]
    \centering
    \includegraphics[width=1.0\linewidth]{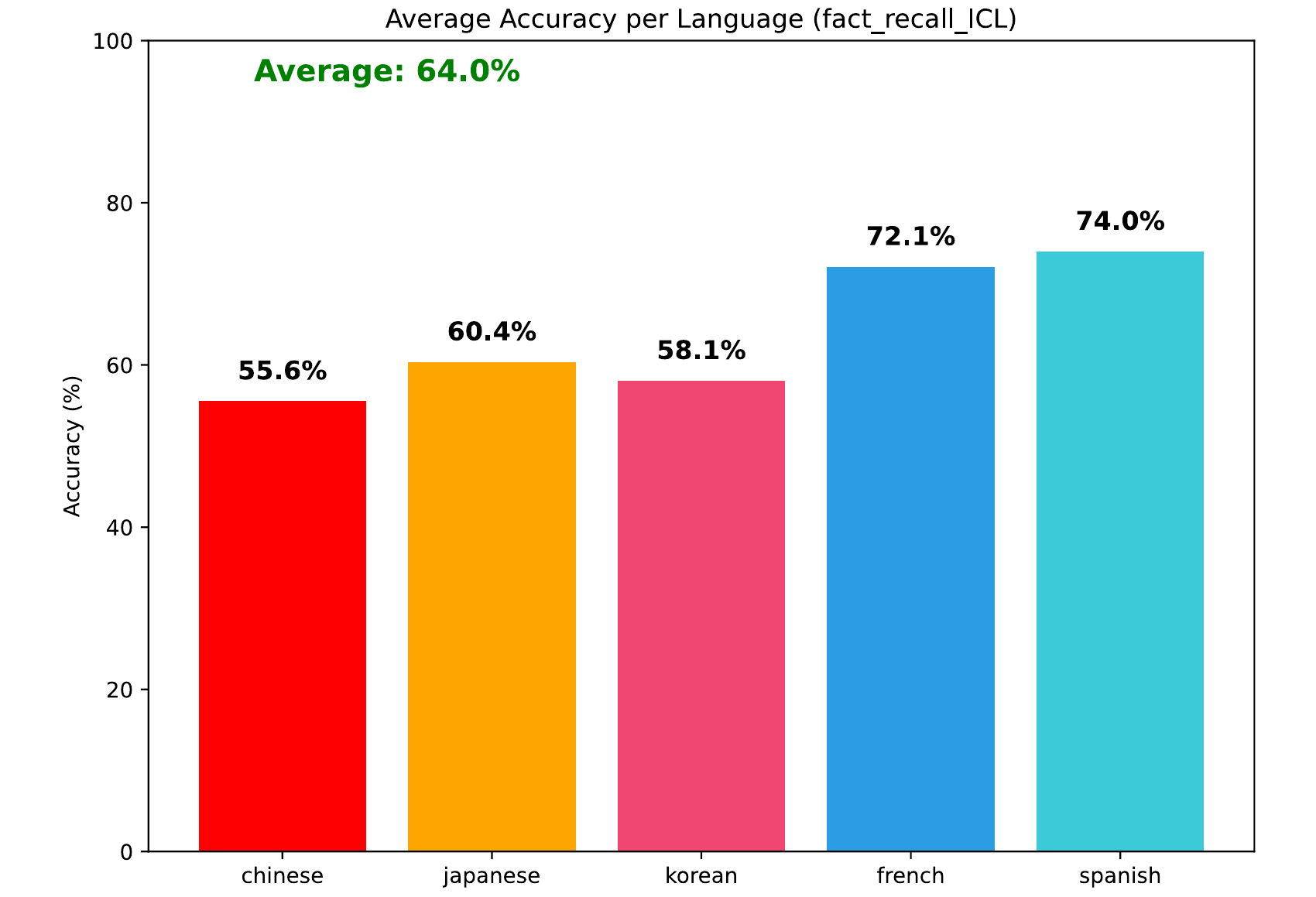}
    \caption{When given 5-shot ICL examples, the fact-recall performance significantly improves for all languages.}
    \label{fig:appendix_ICL_performance}
\end{figure}
\subsection{Relation and Subject Propagation}
\label{sec:propagation_section}
To assess whether—and at which layers—information from the subject and relation positions flows directly to the final token position, we adopt the Attention Knockout method introduced by \citet{geva2023dissectingrecallfactualassociations}, with a slight modification to how relation tokens are defined. 

In the original setup, ~\citet{geva2023dissectingrecallfactualassociations} defines the set of relation positions $R$ as all tokens excluding the subject tokens and the final position. However, since we explicitly annotate relation tokens for each prompt, we use these manually identified indices for $R$ instead. This refinement allows us to more precisely target the positions responsible for encoding the relation. A full list of relation token spans is provided in Table~\ref{tab:relation_words_summary}, and the observed effect also re-validates that these are important positions that carry information that flows to the last token position.

At each layer $\ell$, we block attention from the final token position ($N$) to tokens in $S$ (subject), $R$ (relation), and to itself. This intervention is applied over a sliding window of $k$ layers centered at layer $\ell$, and we measure the resulting change in the model’s prediction probability to evaluate the impact of disrupting this information flow.

We set $k = 6$ following the windowing strategy in \citet{geva2023dissectingrecallfactualassociations}, which corresponds to approximately one-fifth of the total number of model layers and ensures localized but impactful ablations.

\begin{figure}[h]
    \centering
    \includegraphics[width=1\linewidth]{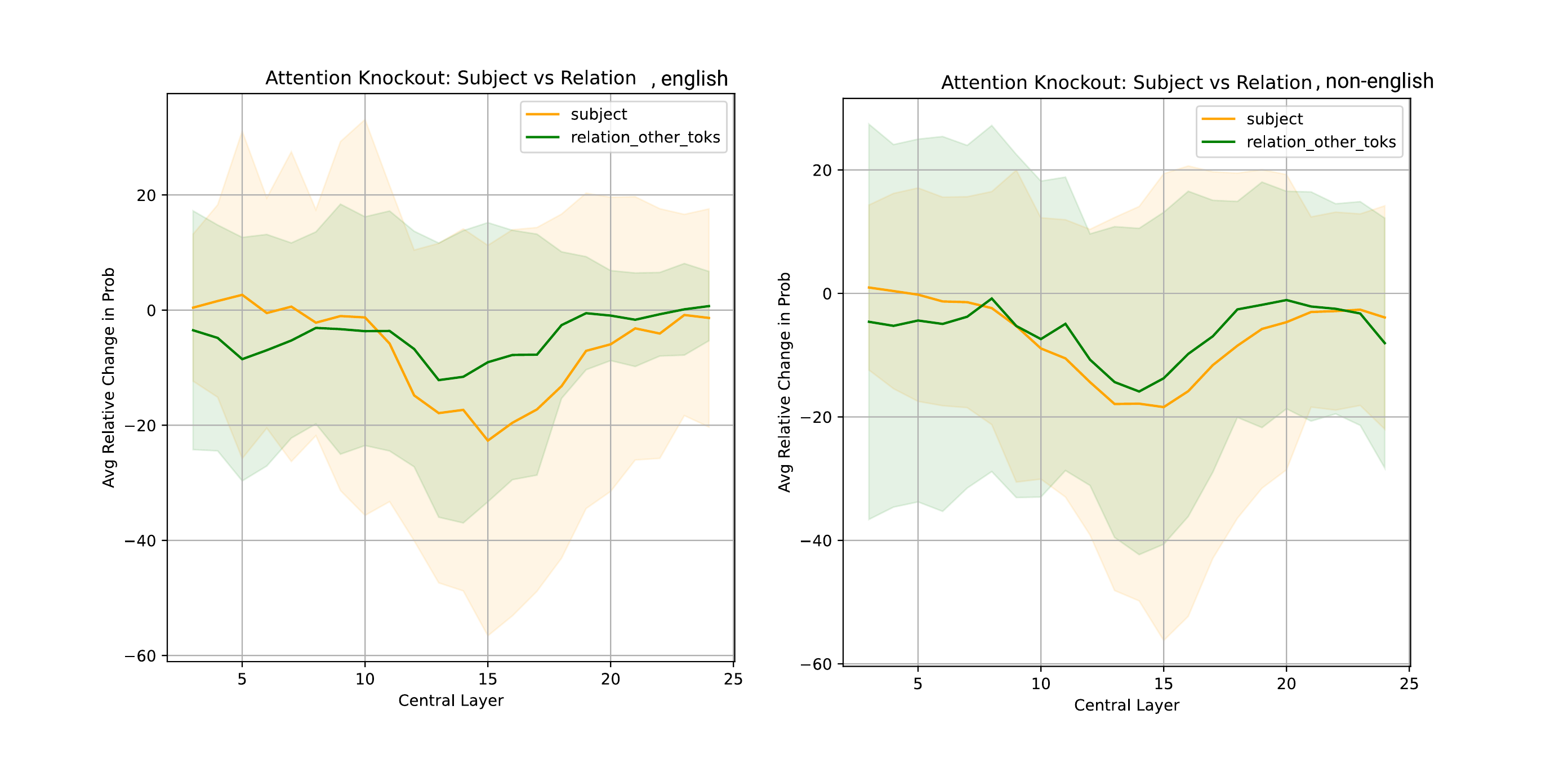}
    \caption{English (left) and Non-English (right) prompts: Blocking attention edges from subject and relation tokens to the last token position causes significant performance drops in layers 10-20, indicating that subject and relation propagation occurs within this layer range.}
    
    \label{fig:experiment_propogation}
\end{figure}

Figure~\ref{fig:experiment_propogation} shows the result for English and non-English cases. The pattern is highly similar: the attention mechanism is responsible for propagating the relation and subject token in layers 10-20. 

\subsection{Extraction Rate}
\label{sec:extraction_section}

\textbf{Experiment Setup} Following \citet{geva2023dissectingrecallfactualassociations}, in order to evaluate whether the model extracts the correct attribute at intermediate layers, we analyze updates to the final token position throughout the model.  At each layer $\ell$, we compute the top-1 token update by projecting the multi-head self-attention or MLP output at the final position to the unembedding matrix. We denote $t^* = \arg\max(p^L_N)$ as the model's final prediction and $t' = \arg\max(E a^\ell_N)$ as the top token from the $\ell$-th layer's update at position $N$ (the final token). 

\citet{geva2023dissectingrecallfactualassociations} observes that in many cases, MLP outputs are simply forwarding the extracted answers from preceding attention layers. To avoid overcounting those as extraction events, we define an \textit{extraction event} as the first layer $\ell$ at which $t' = t^*$. This ensures that we only record the earliest point where the correct English attribute is extracted by either the attention or the MLP pathway.

We compute the \textit{extraction rate} by measuring the frequency of such earliest-match events at each layer. This analysis is conducted independently for both attention and MLP outputs to compare their relative contributions to attribute extraction across the model.

\begin{figure*}[h]
    \centering
    \includegraphics[width=1\linewidth]{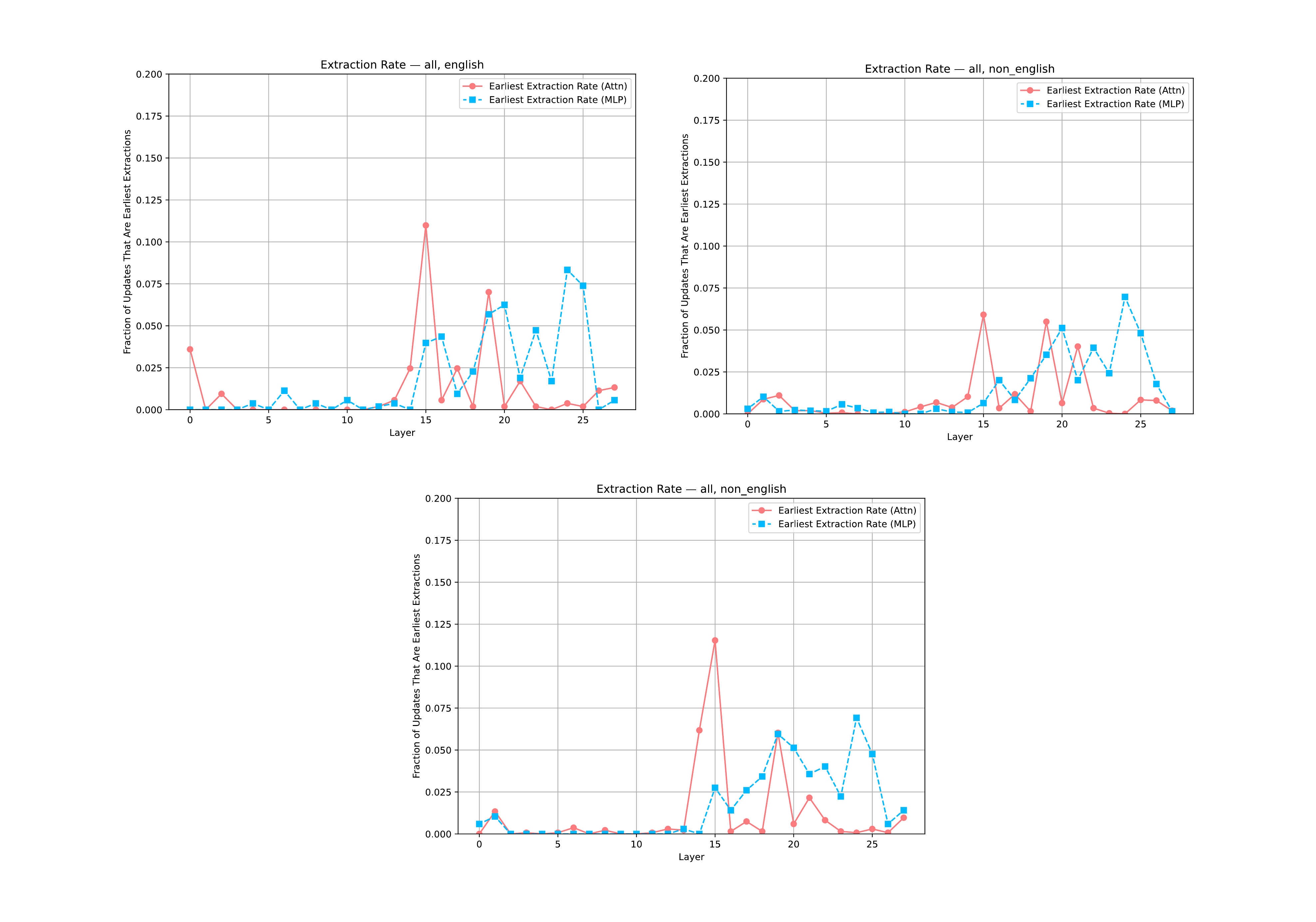}
    \caption{Attribute extraction rate using attention and MLP modules (red and blue respectively) across layers for three conditions (English + Original, Non-English + Original, Non-English + Intervention). Intervention re-activates the attention layers responsible for correct English answer extraction.}
    
    \label{fig:experiment_extraction}
\end{figure*}

Figure~\ref{fig:experiment_extraction} shows the result for the original cases for English and non-English conditions, and non-English after intervention. We observe that for the original English case, attention modules at layer 15 are especially important. However, in non-English cases, it seems that the attention layer 15 components are not extracting out the correct English object enough. Importantly, intervention re-activates the attention layers responsible for correct English answer extraction.

\subsection{English Fact-Recall Attention Heads for Each Relation-Dataset}

\begin{figure*}[h]
    \centering
    \includegraphics[width=1\linewidth]{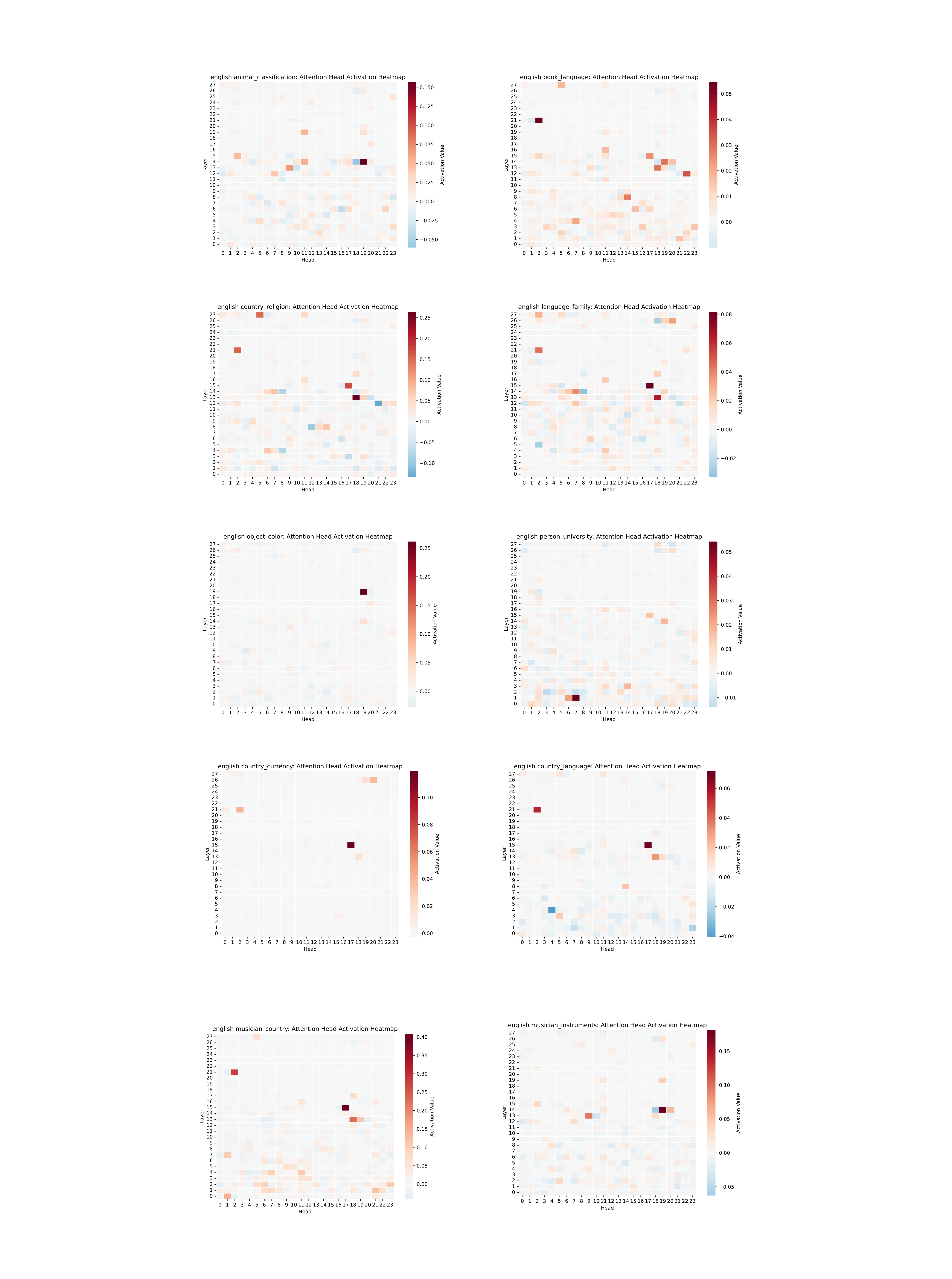}
    \caption{Distinct attention heads are responsible for each English relation-dataset.}    
    \label{fig:english_recall_heads}
\end{figure*}

For answer extraction, \citet{geva2023dissectingrecallfactualassociations}finds that different heads encode subject-answer mappings in their parameters and are specialized for different relation queries. Furthermore, attention heads responsible for the two processes are important and vary across relation-datasets. For instance, through activation patching, we identify that a specific group of heads are most critical for the English \texttt{book\_language} dataset, while a distinct set of heads are crucial for English \texttt{animal\_classification} (Figure~\ref{fig:english_recall_heads}).

\subsection{Ablation of English Fact-Recall Heads}

\begin{figure}[h]
    \centering
    \includegraphics[width=1\linewidth]{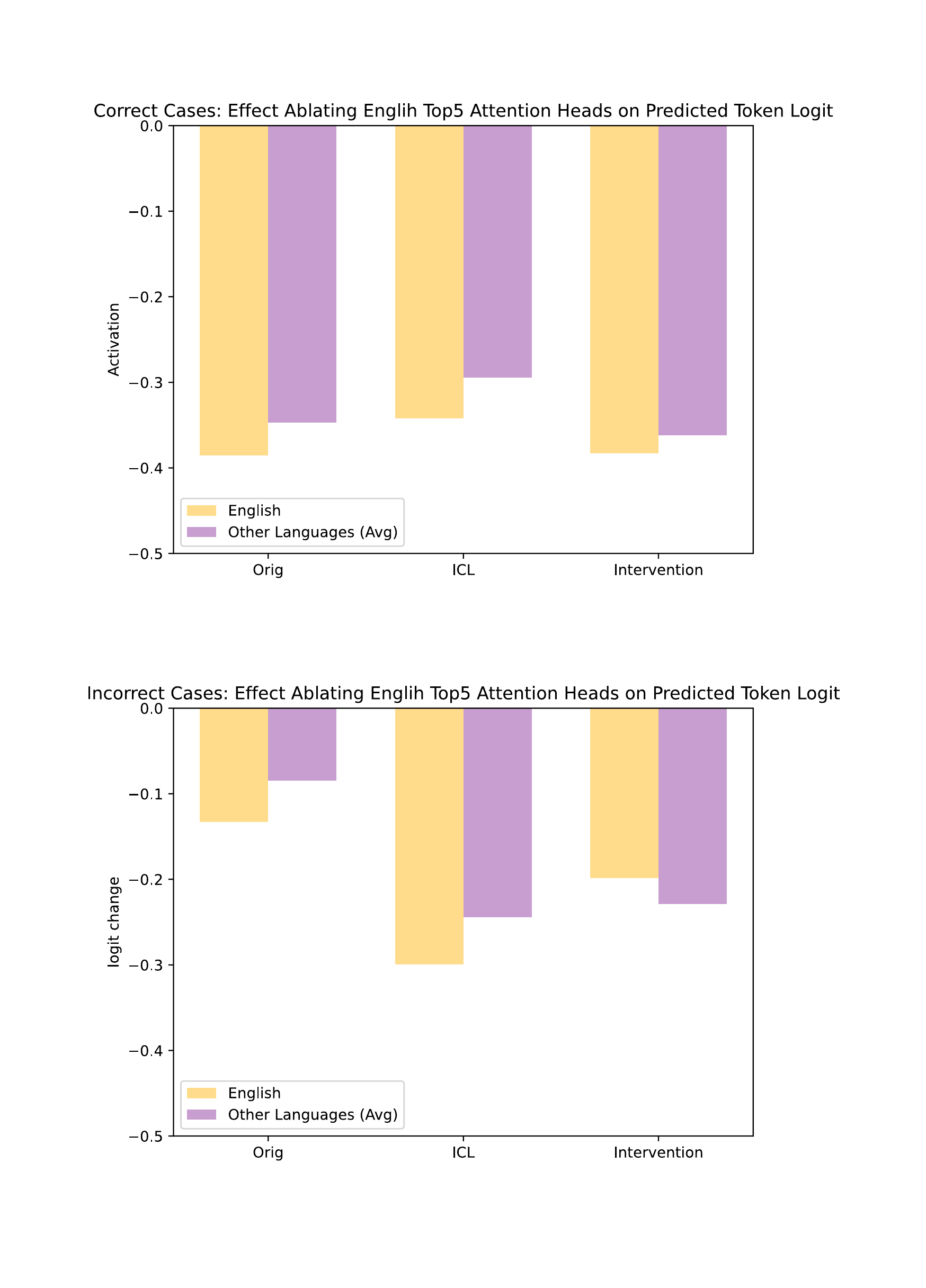}
    \caption{Effect of logits when ablating the top 5 most important dataset-specific English attention heads.}
    \label{fig:head_ablation}
\end{figure}

\begin{figure*}
    \centering
    \includegraphics[width=1\linewidth]{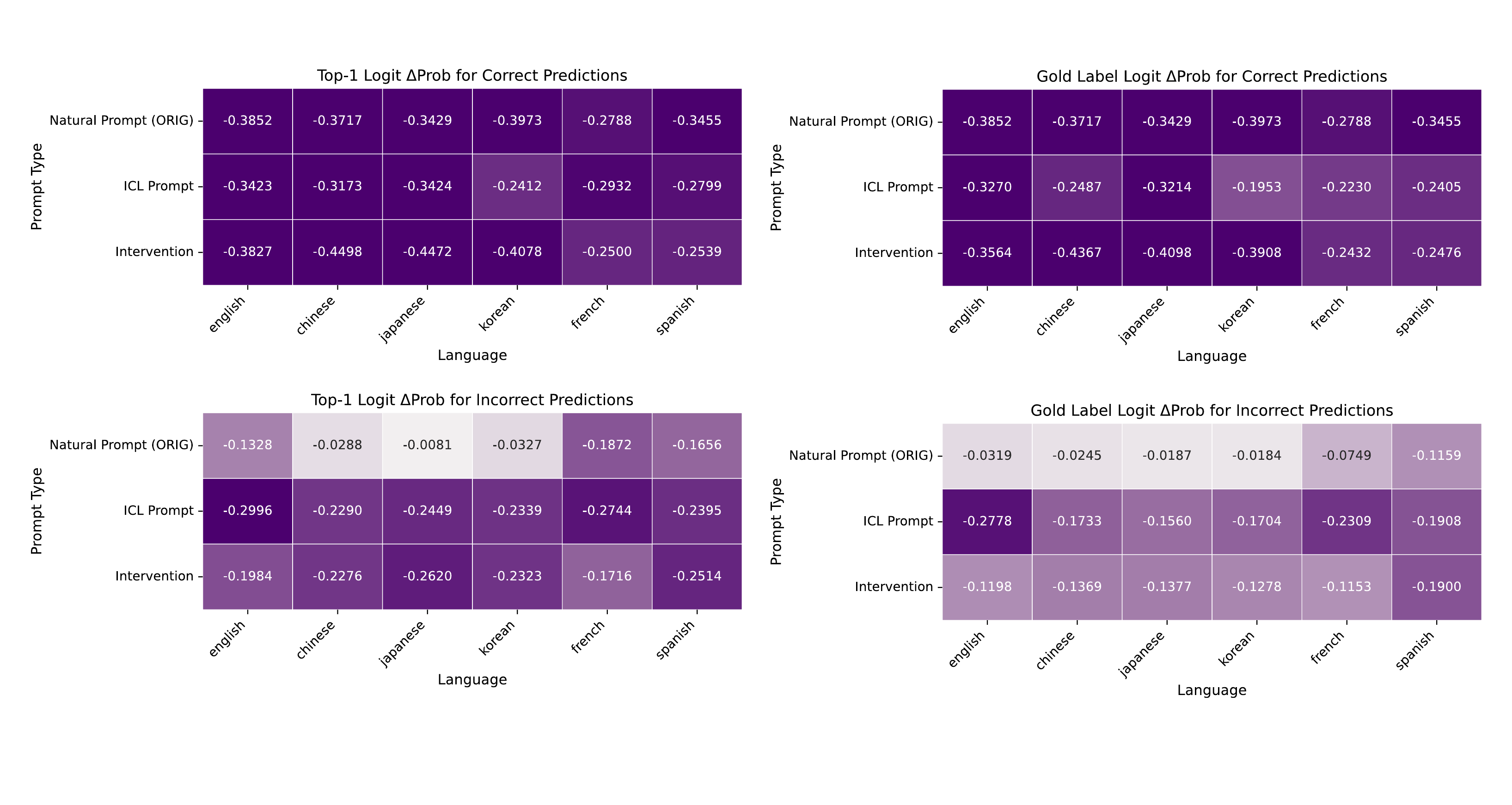}
    \caption{Per-Language Results: The effect of ablating important English Fact-Recall heads in incorrect agnostic cases for each language. English fact-recall heads are not being used to contribute to the model's top1 prediction logits or the model's logits on the label (row 1). The intervention reactivates these heads such that ablating these heads after adding the intervention leads to a significant performance decrease (row 3). The intervention has the same component-level effect as ICL prompting (row 2).}
    \label{fig:top5heads_breakdown}
\end{figure*}

We examine whether the model employs the same significant model heads when given a non-English prompt versus when given an English prompt. To test this hypothesis, we first identify the top 5 most important dataset-specific English attention heads (Figure~\ref{fig:english_recall_heads}) by using activation patching on English correct cases, then ablate these heads to assess their causal role in non-English cases. In Figure~\ref{fig:head_ablation}, we observe in non-English correct cases, we observe significant accuracy drops, which indicates that the model relies on the same English fact-recall components when processing non-English queries. However, in non-English incorrect cases, we observe minimal to no effect, which demonstrates that the model fails to activate these critical English fact-recall pathways. A detailed per-language breakdown of the ablation effect is in Figure~\ref{fig:top5heads_breakdown}.

Importantly, adding non-English ICL examples and adding our dataset-independent and language-independent vectors both reactivate these important attention heads.

\subsection{Implementation Details and Hyperparameters for Recall Vector}
\label{sec:recall_vector_imple_details_hyperparams}
\begin{figure}[h]
    \centering
    \includegraphics[width=1\linewidth]{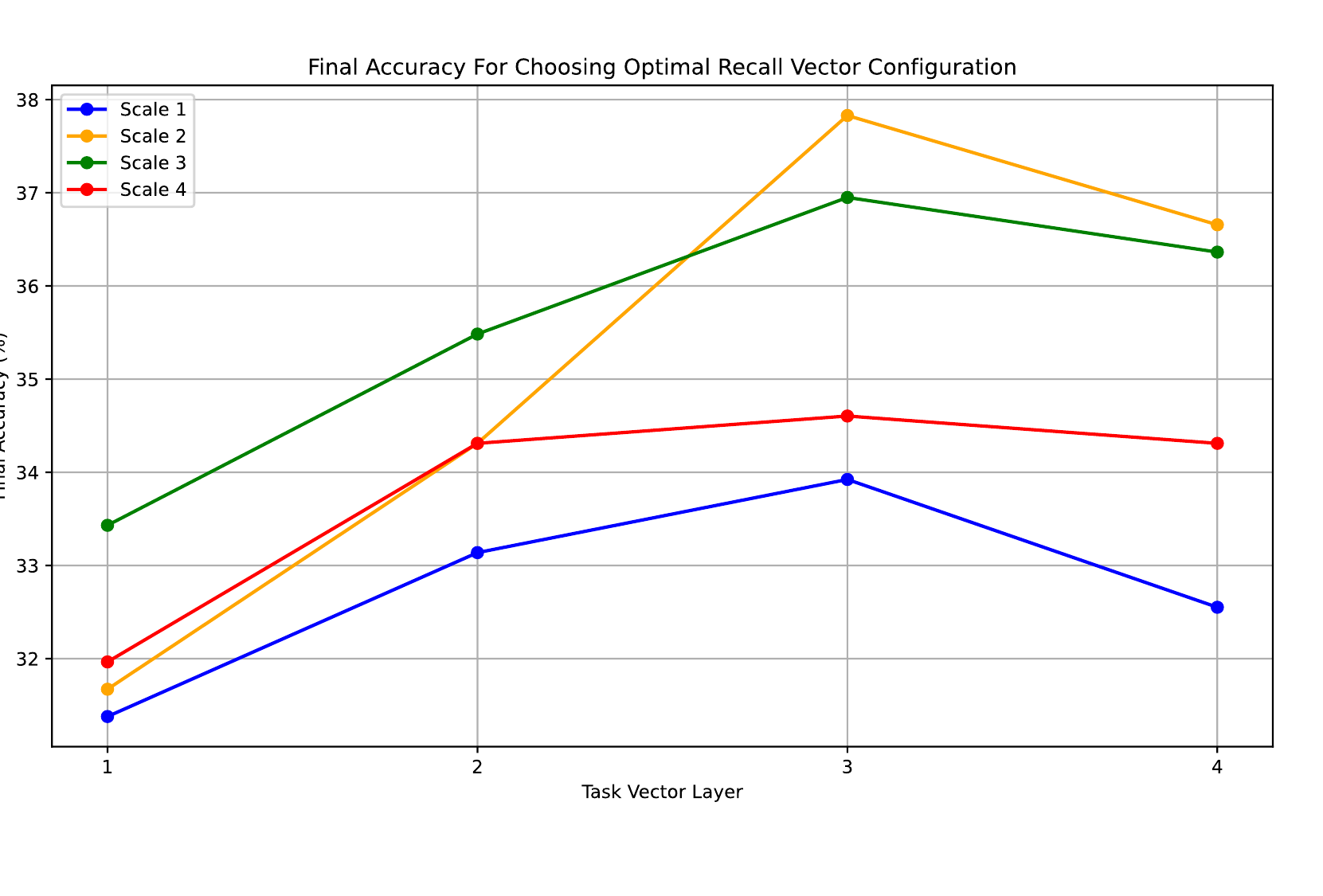}
    \caption{Final Accuracy when extracting the recall task Vector and intervening at various layers and scales.}
    
    \label{fig:recall_opt_hyperparam}
\end{figure}

To identify the optimal configurations for the recall vector, we extract a set of candidate intervention vectors for each intervention layer $\ell \in [L]$ and for a range of scaling factors. Each candidate is evaluated on the held-out validation set to assess its effectiveness. Specifically, we extract the candidate vector from the residual stream at the output of layer $\ell$ and apply the intervention at the beginning of the same layer during inference. Figure~\ref{fig:recall_opt_hyperparam} shows that the best combination is layer 3 with a scaling factor of 2.

\subsection{Intervention Evaluation}
\subsubsection{Optimal Recall Vector Intervention Layer and Scale}
\label{sec:recall_vector_details}
We use the validation dataset to determine the optimal layer and scale for the task vector intervention. Specifically, we inject the task vector at all layers (0-5) and vary the scales (1-5). 

\clearpage
\section{Intervention Evaluation}
\begin{figure*}[h]
    \centering
    \includegraphics[width=1.0\linewidth]{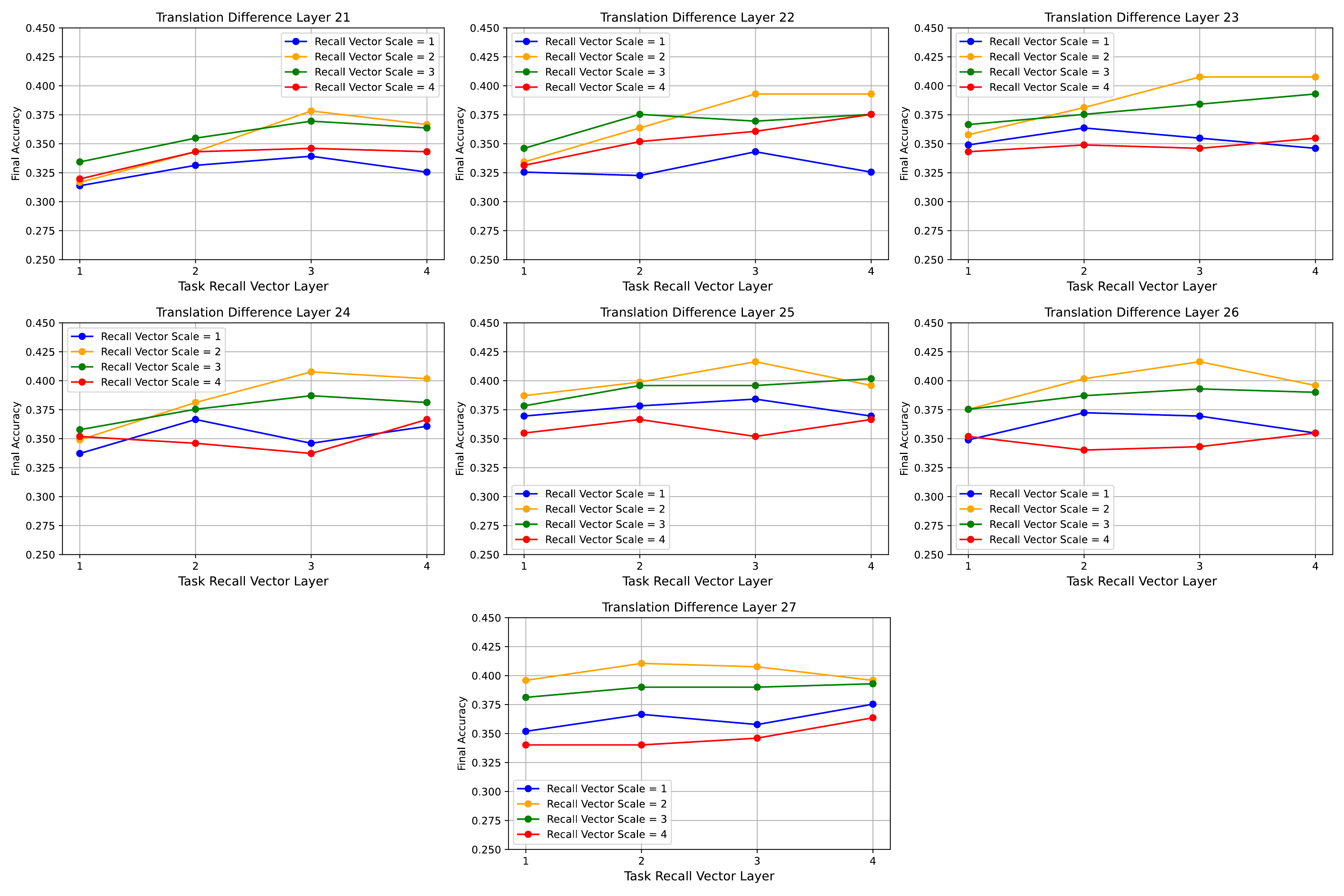}
    \caption{Final Accuracy for intervening using the combined vectors at different layers with different scaling factors for the recall vector on the validation set.}
    \label{fig:combined_opt_param}
\end{figure*}
We perform a grid search over translation difference vectors applied at layers 21 to 26 and recall task vectors from layers 1 to 4, each scaled by factors ranging from 1 to 4, in order to identify the optimal hyperparameter combination. The results are shown in Figure~\ref{fig:combined_opt_param}. We find that applying the translation vector at layer 25 and the recall vector at layer 3, both with a scaling factor of 2, yields the highest validation final accuracy.

\subsection{Intervention Effect Comparison}
\label{sec:appendix_intervention_comparison}

\begin{table}[h]
\centering
\resizebox{1\columnwidth}{!}{%
\begin{tabular}{lrrrr}
\toprule
\textbf{Language}& \textbf{Agn\%} & \textbf{TransAcc\%} & \textbf{Acc\%} \\
\hline
chinese  & 17.93& 0.00  & 2.39  \\
japanese & 11.69& 34.48 & 6.45  \\
korean   & 18.95& 40.43 & 10.48 \\
french   & 53.60& 72.39 & 46.40 \\
spanish  & 45.98& 50.49 & 36.16 \\
english  & 56.05& 81.29 & 61.29 \\
\hline
non-eng  & 29.32& 49.45 & 20.07 \\
\bottomrule
\end{tabular}
}
\caption{Performance Summary (Original)}
\end{table}
\begin{table}[h]
\centering
\resizebox{1\columnwidth}{!}{%
\begin{tabular}{lrrrr}
\toprule
\textbf{Language} & \textbf{Agn\%} & \textbf{TransAcc\%} & \textbf{Acc\%} \\
\hline
chinese   & 23.11& 67.24& 24.70\\
japanese & 24.15& 57.89& 19.49\\
korean   & 23.39& 68.97& 24.19\\
french   & 47.60& 80.67& 47.20 \\
spanish  & 48.80& 63.93& 41.20\\
english & 56.05& 75.54& 58.87\\
\hline
non-eng  & 33.52& 67.74& 31.50\\
\bottomrule
\end{tabular}
}
\caption{Performance Summary (Intervention 1: Translation Difference Vector Intervention)}
\end{table}
\begin{table}[h]
\centering
\resizebox{1\columnwidth}{!}{%
\begin{tabular}{lrrrr}
\toprule
\textbf{Language}& \textbf{Agn\%} & \textbf{TransAcc\%} & \textbf{Acc\%} \\
\hline
chinese   & 41.83 & 59.05 & 36.25 \\
japanese  & 50.00 & 53.23 & 31.45 \\
korean    & 43.55 & 56.48 & 29.03 \\
french    & 56.40 & 70.92 & 46.00 \\
spanish  & 66.40 & 65.66 & 54.00 \\
english   & 68.15 & 89.94 & 72.18 \\
\hline
non-eng   & 51.64 & 61.07 & 39.37 \\
\bottomrule
\end{tabular}
}
\caption{Performance Summary (Intervention 2: Recall Task Vector Intervention).}
\end{table}
\begin{table}[h]
\centering
\resizebox{1\columnwidth}{!}{%
\begin{tabular}{lrrrr}
\toprule
\textbf{Language} & \textbf{Agn\%} & \textbf{TransAcc\%} & \textbf{Acc\%} \\
\hline
chinese   & 47.81 & 61.67 & 41.83 \\
japanese & 53.23 & 55.30 & 33.87 \\
korean    & 47.18 & 52.14 & 29.03 \\
french    & 60.00 & 71.33 & 48.40 \\
spanish   & 66.00 & 66.06 & 52.80 \\
english   & 64.11 & 79.87 & 63.71 \\
\hline
non-eng  & 54.85 & 61.30 & 41.22 \\
\bottomrule
\end{tabular}
}
\caption{Performance Summary (Combined intervention: Translation + Recall Vectors).}
\label{tab:tab_all_interventions}
\end{table}

\subsection{Details of Baseline Experiments}
\label{sec:appendix_baseline_comparison}
To evaluate the impact of interventions on English-centric behavior, we also measure their effect on the original English fact-recall performance. This is important because interventions that boost multilingual performance may introduce competition and degradation in English accuracy. The exception is the \textit{translate-recall-translate} baseline, where this comparison is not meaningful since the prompt and generations are always translated to or from English.

\subsubsection{Translate-recall-translate Baseline}
The translate-recall-translate baseline is a multi-step prompting strategy in which we query the model with three separate prompts sequentially: explicitly instructs the model to translate the question into English, then
conduct the task in English, and then translate the response back to the target language. For each example, we pass intermediate outputs from one step to the next and use the final output for evaluation. Specifically, to account for the translation errors, we count the model as getting the answer correct if one of the first five generated tokens includes the answer token. 

The reason why the translate-reason-translate still has a poor performance in zero-shot fact-recall is because of the accumulation of translation errors. See example failure cases in Table~\ref{tab:translation_failures}.

\begin{table*}[t]
\small
\centering
\begin{tabularx}{\textwidth}{|X|X|X|X|X|}
\hline
\textbf{Failure Type} & \textbf{Original Prompt, attribute} & \textbf{1. Prompt English Translation} & \textbf{2. Fact-Recall Answer} & \textbf{3. Translate Answer back to} \\
\hline

Failed at 1: better translation is “the frog is classified biologically as a” — translating the whole sentence introduces syntax confusion and leads to failures. & 
青蛙在生物学上被分类为一种，\textcolor{darkgreen}{两栖动物} & 
The frog is classified as a biological species & 
, which means that it is a group of organisms that share a common ancestor and are able to reproduce with each other. & 
\textcolor{red}{非常不一样的语言。} \\
\hline

Failed at 1: mistranslates “the book’s Notre-Dame de Paris was originally written in” — instead refers to the cathedral’s architecture. & 
巴黎圣母院最初编写时使用的语言为，\textcolor{darkgreen}{法语} & 
The original language used in Notre Dame Cathedral was & 
Latin. The Latin language was the language of the Roman Catholic Church. & 
\textcolor{red}{拉丁语} \\
\hline

Failed at 2 and 3. & 
毛利语所属的语系是，\textcolor{darkgreen}{南岛语系} & 
The language family to which the Moli language belongs is & 
The Moli language family. The Moli language is spoken in the Moli language region. & 
The \textcolor{red}{Moli language} family. The Moli language is spoken in the Moli language region. \\
\hline

Failed at 3: correct in steps 1 and 2 but failed to translate the correct language-specific answer back. & 
希伯来语所属的语系是，\textcolor{darkgreen}{亚非语系} & 
The language family to which Hebrew belongs is & 
Called Afro-Asiatic. It is the largest language family in Africa and the second largest in the Middle East. & 
叫做\textcolor{red}{非洲-亚洲语族}。它是非洲最大的语言家族，中东的第二大。 \\
\hline
\end{tabularx}
\caption{Examples of failure cases from the translate-recall-translate baseline experiment. \textcolor{darkgreen}{Green} highlights correct answers; \textcolor{red}{red} indicates incorrect outputs.}
\label{tab:translation_failures}
\end{table*}

\subsubsection{Fine-tuning Baseline}

We split our data into train-val-test subsets according to a 40-10-50 ratio. Using 2 NVIDIA L40S GPUs, we finetune Llama-3.2-3B on the train subset for 30 epochs using the AdamW optimizer with a learning rate of $1 \times 10^{-5}$ and pick the best checkpoint using the validation performance. All training and inference runs are conducted using the transformer library\footnote{\url{https://huggingface.co/docs/transformers/index}}.
\clearpage
\section{Evaluation with a different splitting strategy}
\label{sec:across_dataset_performance}
\begin{figure}[htbp]
    \centering
    \includegraphics[width=\linewidth]{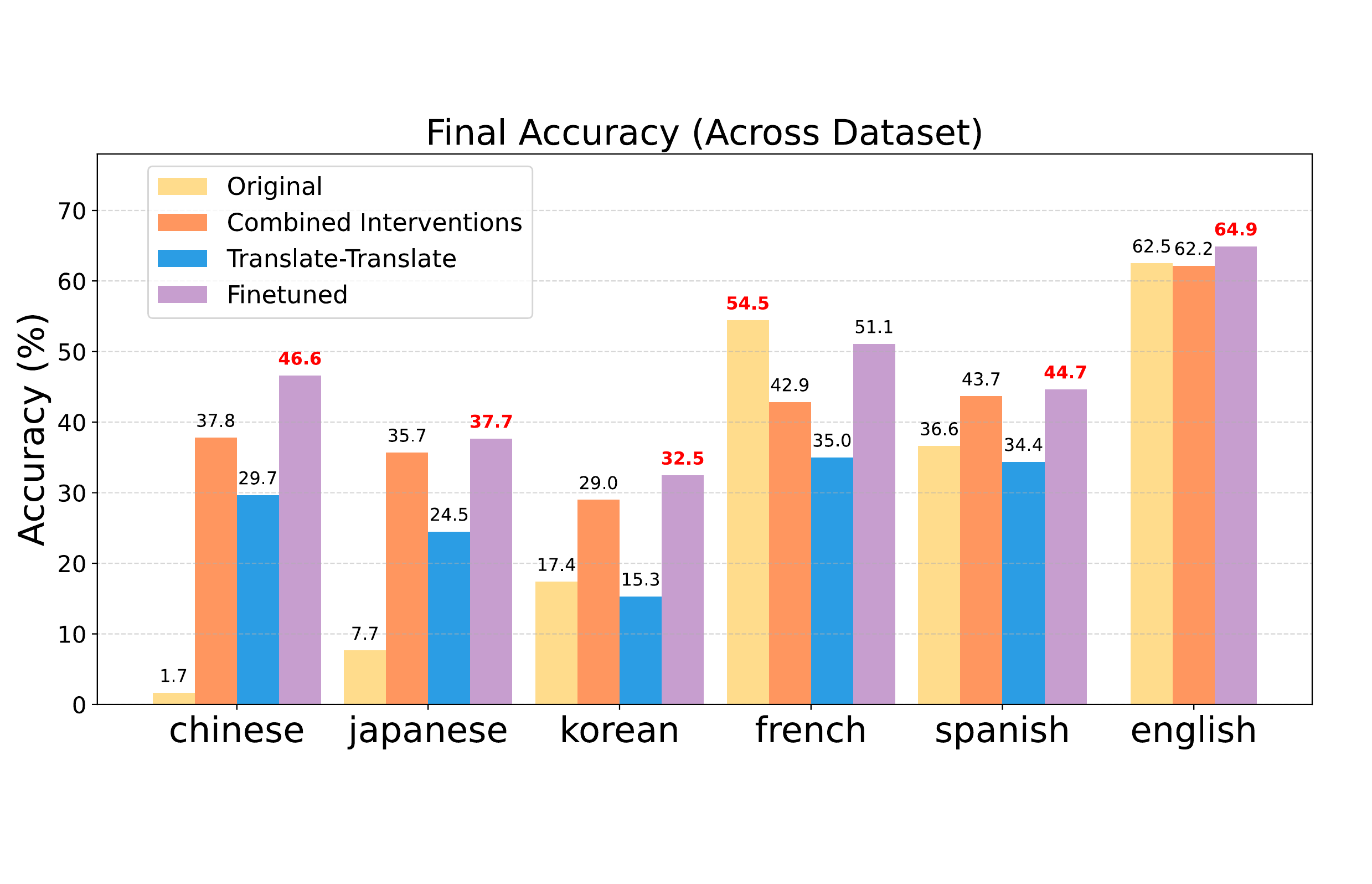}
    \caption{Intervention performance compared to baselines across test sets, averaged over three random seeds. This shows splits across relation datasets, the right shows splits within each relation-specific dataset. }
    \label{fig:baseline_across_dataset}
\end{figure}

We not only evaluate the intervention's effect on the within-relation train-val-test split, where we split each relation-dataset into train, val, test subsets independently (Figure~\ref{fig:final_performance}(b)), but we also evaluate across-relation split: we train on a subset of relation-datasets and evaluate on held-out, unseen relations to assess the model's ability to generalize factual recall beyond previously encountered answer types. Figure~\ref{fig:baseline_across_dataset} shows that our intervention is significantly better than translate-translate baseline and is competitive with fine-tuning when generalizing to new relation-datasets.

%%%%%%%%%%%%%%%%%%%%%%%%%%%%%%%%%%%%%%%%%%%%%%%%%%%%%%%%%%%%

\end{CJK*}

\end{document}